\documentclass[letterpaper]{article} % DO NOT CHANGE THIS
\usepackage[preprint]{aaai2027}
\usepackage[hyphens]{url}  % DO NOT CHANGE THIS
\usepackage{graphicx} % DO NOT CHANGE THIS
\usepackage{natbib}  % DO NOT CHANGE THIS AND DO NOT ADD ANY OPTIONS TO IT
\usepackage{caption} % DO NOT CHANGE THIS AND DO NOT ADD ANY OPTIONS TO IT
\usepackage{algorithm}
\usepackage{algorithmic}
\usepackage{amsmath}
\usepackage{tabularx}
\usepackage{amssymb}
\usepackage{array}
\usepackage{longtable}
\usepackage{booktabs}
\usepackage{multirow}

\usepackage{newfloat}
\usepackage{listings}
\DeclareCaptionStyle{ruled}{labelfont=normalfont,labelsep=colon,strut=off} % DO NOT CHANGE THIS
\floatstyle{ruled}
\newfloat{listing}{tb}{lst}{}
\floatname{listing}{Listing}

\usepackage{booktabs}

\title{SERL-SQL: Selective Hindsight Distillation for Text-to-SQL Reinforcement Agentic Learning}
\author{
    Tao Liu,\textsuperscript{\rm 1}\
    Tao Feng,\textsuperscript{\rm 1}
    Xiangheng Li,\textsuperscript{\rm 2}
    Jinwang Song,\textsuperscript{\rm 2}
    Yifan Li,\textsuperscript{\rm 2}
    Xiaoqing Cheng,\textsuperscript{\rm 2}
    Dixuan Zhang,\textsuperscript{\rm 2}
    Siquan Li,\textsuperscript{\rm 2}
    Lin Lan,\textsuperscript{\rm 2}
    Hongying Zan,\textsuperscript{\rm 2}
    Kunli Zhang,\textsuperscript{\rm 2}
    Chao Wu\textsuperscript{\rm 1}\corresponding\\
}
\affiliations{
    \textsuperscript{\rm 1}Zhejiang University, College of Artificial Intelligence\\
    \textsuperscript{\rm 2}Zhengzhou University, School of Computer Science of Artificial Intelligence\\
}

\begin{document}

\maketitle

\begin{abstract}
Recent Text-to-SQL systems increasingly rely on multi-turn interaction, execution feedback, and reinforcement learning. However, most existing methods use execution correctness only as a trajectory-level reward, which provides limited guidance for identifying the SQL decisions responsible for success or failure. We propose \textbf{SERL-SQL}, a selective execution-grounded reinforcement learning framework for multi-turn Text-to-SQL agents. \textbf{SERL-SQL} samples on-policy SQL interaction trajectories and uses a training-only teacher to re-score student actions with execution feedback. The resulting teacher--student likelihood gap is converted into bounded, masked weights that reweight GRPO advantages only on SQL and tool-action tokens. In this way, task rewards preserve the optimization direction, while execution hindsight provides localized credit assignment. Experiments on BIRD, Spider, and cross-domain benchmarks show that \textbf{SERL-SQL} achieves competitive performance, reaching 76.56\% execution accuracy on BIRD-Dev and 89.92\% on Spider-Test. Moreover, our reward-based selection strategy closely approaches the oracle Best-of-N upper bound and consistently outperforms consistency-based selection, showing that \textbf{SERL-SQL} produces high-quality candidates that can be reliably identified by lightweight execution-grounded rewards.
\end{abstract}

\begin{links}
    \link{Code}{https://github.com/Ffunkytao/SERL-SQL}
    % \link{Datasets}{https://aaai.org/example/datasets}
    % \link{Extended version}{https://aaai.org/example/extended-version}
\end{links}

\section{Introduction}
Text-to-SQL translates natural language questions into executable SQL queries over relational databases. Recent Text-to-SQL research has increasingly shifted from in-context learning~\cite{wang2023mac,gao2024xiyansql,pourreza2023dinsql,dailsql} generation to agentic database interaction. Instead of producing a single query directly, modern SQL agents~\cite{pourreza2024chasesql,talaei2024chess,li2025deepeyesql} decompose the process into schema grounding, SQL generation, execution, verification, revision, and selection, enabling models to use database feedback throughout the reasoning process. Representative systems such as Agentar-Scale-SQL~\cite{wang2025agentarscalesql}, Arctic-Text2SQL-R1~\cite{yao2026arctictext2sqlr1simplerewardsstrong} and MARS-SQL~\cite{yang2025marssql} further demonstrate the effectiveness of scaling agentic interaction and reinforcement learning for robust Text-to-SQL performance. These advances suggest that realistic Text-to-SQL is no longer merely a sequence generation problem, but an interactive decision-making process grounded in execution.

\begin{figure}[t]
  \centering
  \includegraphics[width=\linewidth]{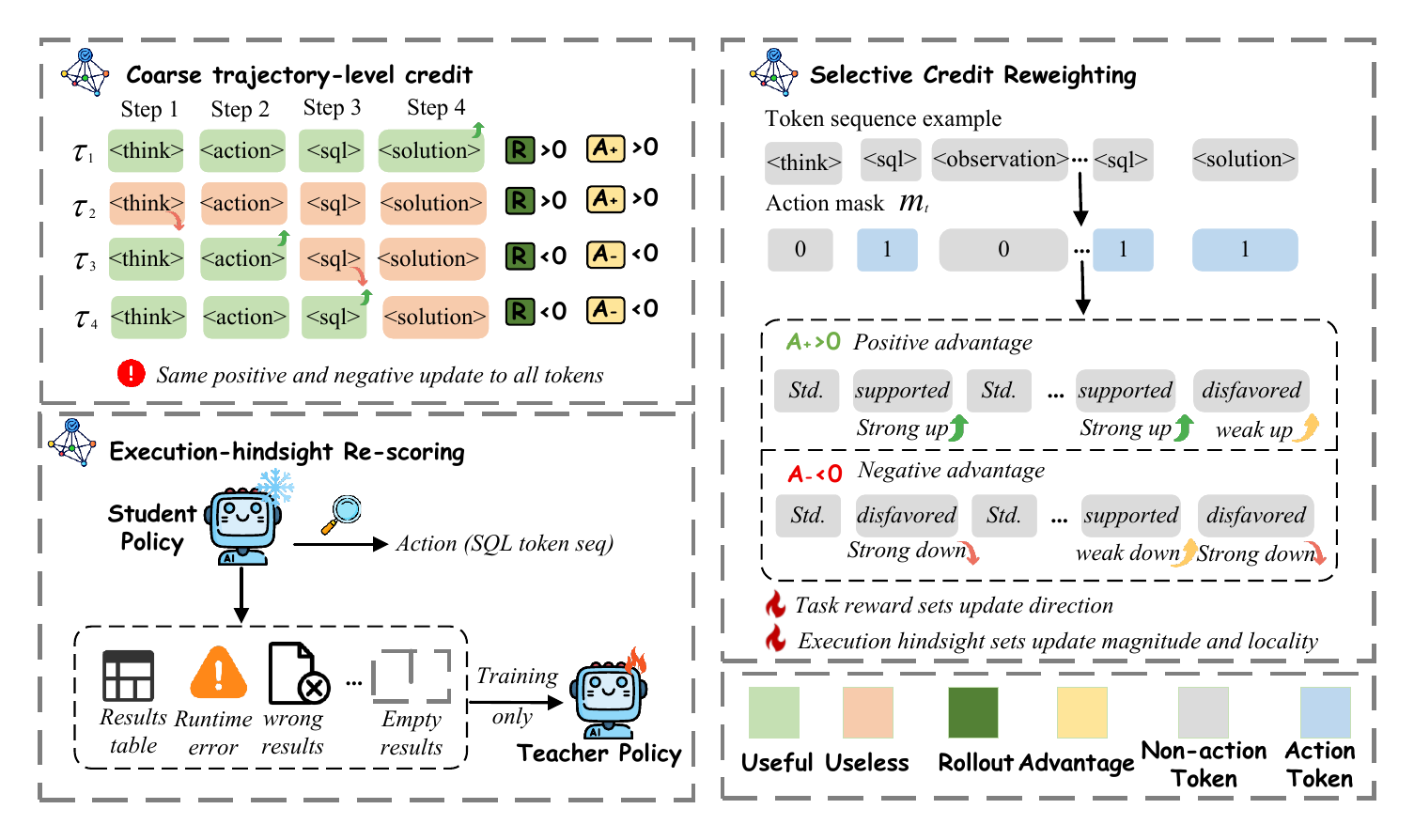}
  \caption{SERL-SQL uses execution hindsight to convert coarse trajectory-level GRPO rewards into selective, reward-aligned credit reweighting for SQL action tokens.}
  \label{fig1:motiavation}
\end{figure}

Existing agentic Text-to-SQL systems have explored several ways to improve SQL generation through execution-aware training. Search-based and progressive RL methods improve candidate exploration or reward shaping during policy optimization~\cite{ma2025sqlr1trainingnaturallanguage,zhang2026progresssqlimprovingreinforcementlearning}, while SQL-Trail~\cite{hua-etal-2026-sql} use interleaved execution feedback to guide iterative refinement and trajectory evaluation. But they lack real interaction with database, further many works~\cite{guo-etal-2026-mtsql,dai-etal-2026-reex,hua-etal-2026-sql,li-etal-2026-sql} exploit database feedback for correction, validation, or candidate selection. And FineStep~\cite{dai2026stepcountssteplevelcredit} show the importance of credit assignment by moving toward finer supervision by assigning credit at the step level for tool-integrated Text-to-SQL. However, these coarse signals cannot localize the SQL decisions responsible for errors, causing credit to diffuse beyond the executable actions that determine correctness.

Our key insight is to use execution feedback not only as an outcome reward, but also as training-time hindsight for selective credit assignment. Execution errors, empty results, or incorrect outputs provide privileged evidence about which SQL decisions may be responsible for failure. Instead of exposing such information to the deployed policy or distilling it uniformly, \textbf{SERL-SQL} uses a synchronized teacher to re-score the student's on-policy actions under the hindsight context. The resulting teacher--student log-probability gap is converted into a bounded GRPO reweighting factor~\cite{shao2024deepseekmath}, applied only to SQL-relevant and tool-action tokens through a selective action mask. Thus, task rewards preserve the optimization direction, while execution hindsight controls the locality and strength of updates. At inference time, the teacher is removed, and the learned policy operates as a standalone SQL agent.

We evaluate \textbf{SERL-SQL} on BIRD, a realistic large-scale database-grounded Text-to-SQL benchmark~\cite{li2023bird}. Our current best model achieves 76.56\% execution accuracy with reward-based selection over sampled trajectories, closely approaching the oracle best-of-$N$ result of 77.84\%. This small gap suggests that \textbf{SERL-SQL} learns policies whose successful trajectories can be identified by lightweight execution-grounded signals.
\noindent In summary, our contributions are as follows:
\begin{itemize}
    \item We formulate execution-grounded on-policy distillation for Text-to-SQL, where database feedback is used as teacher-side privileged context to produce bounded, selective advantage reweighting over SQL-relevant and tool-action-relevant tokens.
    \item We validate \textsc{SERL-SQL} on BIRD and Spider, where it achieves 76.56\% execution accuracy on BIRD-Dev and 89.92\% on Spider-Test. Moreover, reward-based selection closely approaches the oracle Best-of-N upper bound and outperforms consistency-based selection, indicating that SERL-SQL produces high-quality candidates identifiable by execution-grounded rewards.
\end{itemize}

\begin{figure*}[h]
  \centering
  \includegraphics[width=\linewidth]{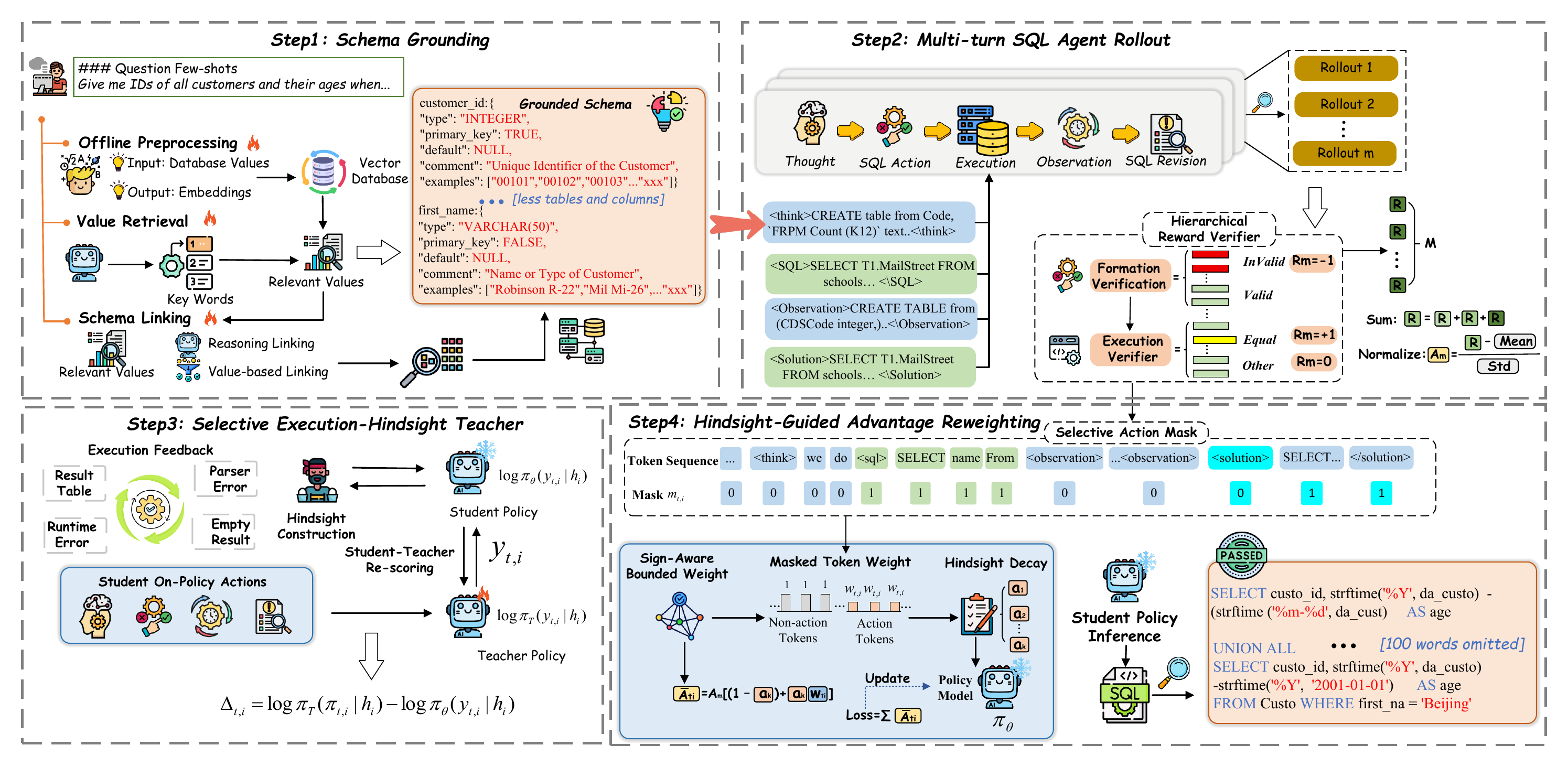}
  \caption{Overview of the SERL-SQL pipeline. (1) schema grounding; (2) multi-turn SQL agent rollout with hierarchical reward verification; (3) selective execution-hindsight re-scoring; and (4) hindsight-guided advantage reweighting.}
  \label{fig:mainpipline}
\end{figure*}

\section{Preliminaries}

\subsection{Problem Formulation}

Given a natural language question $q$, a database schema $S$, and optional external knowledge $e$~\cite{10.1145/3737873,liu2025surveytexttosqlerallms}, the agent generates an executable SQL query $y$ whose result should match the gold answer. We formulate Text-to-SQL as a multi-turn interaction process. At turn $t$, the policy observes state $s_t$, produces action $a_t$, and receives database observation $o_t$. A trajectory is defined as
\begin{equation}
\tau = (s_0, a_0, o_0, \ldots, s_T, a_T, o_T).
\label{eq:trajectory}
\end{equation}
The final SQL is extracted as $y=\mathrm{Etsql}(\tau)$ and evaluated by execution accuracy~\cite{hong2025nextgenerationdatabaseinterfacessurvey}. For $N$ trajectories sampled from the current policy $\pi_\theta$, the environment returns trajectory-level rewards ${R^n}_{n=1}^{N}$ based on execution correctness.
% Although database observations are informative, trajectory-level rewards provide coarse credit assignment. \textsc{SERL-SQL} instead uses execution hindsight for selective token-level updates.

\subsection{Group Relative Policy Optimization}

GRPO optimizes an on-policy group of trajectories without training an additional value model~\cite{shao2024deepseekmath}. For a given input, let $\{\tau^n\}_{n=1}^{N}$ be trajectories sampled from the old policy $\pi_{\theta_{\mathrm{old}}}$ with rewards $\{R^n\}_{n=1}^{N}$. The group-relative advantage of trajectory $\tau^n$ is
\begin{equation}
A^n =
\frac{
R^n-\operatorname{mean}_{m}(R^m)
}{
\operatorname{std}_{m}(R^m)+\epsilon_A
}.
\label{eq:grpo_advantage}
\end{equation}
In standard trajectory-level GRPO, this reward-derived advantage is assigned to every action step in the trajectory, i.e., $A_t=A^n$. For token $y_{t,i}$, the policy ratio is
\begin{equation}
r_{t,i}(\theta)=
\frac{
\pi_\theta(y_{t,i}\mid h_t,y_{t,<i})
}{
\pi_{\theta_{\mathrm{old}}}(y_{t,i}\mid h_t,y_{t,<i})
},
\label{eq:policy_ratio}
\end{equation}
where $h_t$ denotes the interaction history before action $a_t$. For any token-level advantage $B_{t,i}$, we define the clipped GRPO
surrogate as
\begin{equation}
\small
\begin{aligned}
\ell_{\mathrm{GRPO}}(\theta;B_{t,i})
=
\min\Big(
&r_{t,i}(\theta)B_{t,i},\\
&\operatorname{clip}\!\left(
r_{t,i}(\theta),1-\epsilon_p,1+\epsilon_p
\right)B_{t,i}
\Big).
\end{aligned}
\label{eq:grpo_surrogate}
\end{equation}
The standard GRPO objective is then
\begin{equation}
\small
\mathcal{L}_{\mathrm{GRPO}}(\theta)
=
-\mathbb{E}_{\tau}
\left[
\sum_{t,i}
\ell_{\mathrm{GRPO}}(\theta;A_t)
\right]
+
\beta\,\mathrm{KL}\!\left(
\pi_\theta\|\pi_{\mathrm{ref}}
\right).
\label{eq:grpo_objective}
\end{equation}
Although this objective provides a stable reward-aligned update direction, its credit assignment remains coarse: decisive SQL actions, reasoning tokens, and formatting tokens receive the same trajectory-level advantage.

\subsection{On-Policy Distillation}
On-policy distillation~\cite{agarwal2024onpolicy,zhao2026selfdistilledreasoner} provides a complementary token-level training signal. A teacher policy $\pi_T$ evaluates trajectories sampled from the current student policy and supplies dense supervision:
\begin{equation}
\mathcal{L}_{\mathrm{OPD}}
=
\sum_{t,i}
\mathrm{KL}\!\left[
\pi_T\!\left(
\cdot \mid h_t,y_{t,<i},c_t^{\mathrm{ex}}
\right)
\,\middle\|\,
\pi_\theta\!\left(
\cdot \mid h_t,y_{t,<i}
\right)
\right],
\label{eq:opd_objective}
\end{equation}
where $c_t^{\mathrm{ex}}$ denotes teacher-only execution hindsight, such as query results, runtime errors, parser errors, or empty outputs. Directly imitating such a hindsight-conditioned teacher may introduce information leakage, teacher bias, and optimization instability. \textsc{SERL-SQL} therefore uses this signal only for bounded, selective credit reweighting over SQL and tool-action tokens.

% ============================================================
% Methodology
% ============================================================

\section{Methodology}

\textsc{SERL-SQL} consists of four stages, as shown in Figure~\ref{fig:mainpipline}. It first grounds the full schema into a compact database context, then samples multi-turn SQL trajectories through iterative generation, execution, and revision. A training-only teacher re-scores the student’s on-policy actions with execution hindsight, and the resulting teacher--student likelihood gap is used to selectively reweight GRPO advantages over executable action tokens. After training, the teacher is removed and the student policy is deployed independently.

\subsection{Schema Grounding}
\label{sec:schema_grounding}

Given a natural language question $q$ and a database schema $S$, the schema grounding agent identifies a compact schema subset $S'\subseteq S$ containing the tables, columns, and database values
required to answer the question. As illustrated in Figure~\ref{fig:mainpipline}, the grounding process consists of offline value indexing, question-aware value retrieval, and hybrid schema linking. Let
$S=\{(T_j,C_j)\}_{j=1}^{M}$
% \[
% S=\{(T_j,C_j)\}_{j=1}^{M}
% \]
denote the full schema, where $T_j$ is a table and $C_j$ is its set of columns. During offline preprocessing, representative values from each column are encoded by an embedding model $E_{\eta}$ and stored in a column-aware vector index:
\begin{equation}
\mathcal{I}
=
\left\{
(T_j,c,v,E_{\eta}(v))
\mid
T_j\in S,\;
c\in C_j,\;
v\in\mathcal{V}_{j,c}
\right\},
\label{eq:value_index}
\end{equation}
where $\mathcal{V}_{j,c}$ denotes the values associated with column $c$. This index is constructed once for each database and reused across different questions.

At query time, keywords and entity mentions $\mathcal{K}_q$ are extracted from the question and matched against the vector index. The top-ranked values and their source columns form the retrieved value
evidence:
\begin{equation}
\mathcal{V}_{q}
=
\operatorname{TopK}_{(T_j,c,v)\in\mathcal{I}}
\max_{k\in\mathcal{K}_{q}}
\operatorname{sim}
\bigl(E_{\eta}(k),E_{\eta}(v)\bigr).
\label{eq:value_retrieval}
\end{equation}
Such evidence links question expressions to concrete database contents, especially when the question and stored values use different surface forms.

We then combine reasoning-based linking and value-based linking. Reasoning-based linking identifies schema elements required by the semantic structure of the question, such as projected columns, filtering
conditions, join keys, and aggregation attributes. Value-based linking selects columns supported by the retrieved database values. The grounding module therefore produces:
\begin{equation}
S'
=
G_{\phi}(q,S,\mathcal{V}_{q})
=
\left\{
(T_j,C'_j)
\mid
T_j\in S,\;
C'_j\subseteq C_j
\right\}.
\label{eq:schema_grounding}
\end{equation}

Each retained column is serialized with its data type, key information, description, and representative values. The resulting grounded schema provides a compact and value-aware database context shared by all
subsequent policy rollouts, reducing irrelevant schema information while preserving the evidence required for SQL generation.

\subsection{Multi-turn SQL Agent Rollout}
\label{sec:multi_turn_rollout}

Conditioned on the question $q$, the grounded schema $S'$, and optional external knowledge $e$, the student policy $\pi_\theta$ interacts with the database through iterative SQL generation, execution, and revision. Rather than directly producing a final query, the policy follows a think, act and observe process, allowing it to inspect execution feedback and revise previous SQL decisions.We formulate this interaction as a process $\mathcal{M}=(\mathcal{S}_{\mathrm{int}},\mathcal{A}, \mathcal{P},\mathcal{R})$.

\paragraph{State Space.}
At turn $t$, the state contains the task input and the complete interaction history, $s_t =\left(q,S',e,h_t\right)$,$h_t=(a_0,o_0,\ldots,a_{t-1},o_{t-1})$. Thus, the policy can condition its next decision on both the original database context and all previous execution feedback.

\paragraph{Action Space.}
The student samples an action as $a_t = \pi_\theta(\cdot\mid s_t)$.
Each action contains a reasoning span followed by either an intermediate SQL action or a terminal solution. Intermediate SQL actions may construct a new query or revise a previous one. When $a_t$ contains an executable SQL query, the database returns an observation $o_t$, such as a result table, runtime error, parser error, or empty result. The environment transition is therefore determined by database execution 
$
s_{t+1}
=
\mathcal{P}(s_t,a_t,o_t).
\label{eq:rollout_transition}
$

the $n$-th sampled trajectory is written as
\begin{equation}
\tau^n =
(s_0,a_0,o_0,\ldots,s_T,a_T,o_T).
\end{equation}
The final SQL query is extracted from the terminal solution:
\begin{equation}
y^n=\operatorname{Etsql}(\tau^n).
\label{eq:extract_sql}
\end{equation}

\paragraph{Trajectory Reward.}
For each input, we sample a group of $N$ on-policy trajectories $\{\tau^n\}_{n=1}^{N}$. A hierarchical verifier first checks whether
the trajectory satisfies the required interaction format and whether a final SQL query can be extracted. It then compares the execution result
of the predicted SQL with that of the gold query. The trajectory reward is defined as
\begin{equation}
R^n =
\mathcal{R}_{\mathrm{ex}}(\tau^n)
=
\begin{cases}
+1, & \text{if $\tau^n$ valid and $y^n$ correct},\\
0,  & \text{if $\tau^n$ valid but $y^n$ incorrect},\\
-1, & \text{if $\tau^n$ invalid}.
\end{cases}
\label{eq:execution_reward}
\end{equation}
This sparse reward determines the group-relative trajectory advantage used by GRPO.

\begin{figure}[t]
  \centering
  \includegraphics[width=\linewidth]{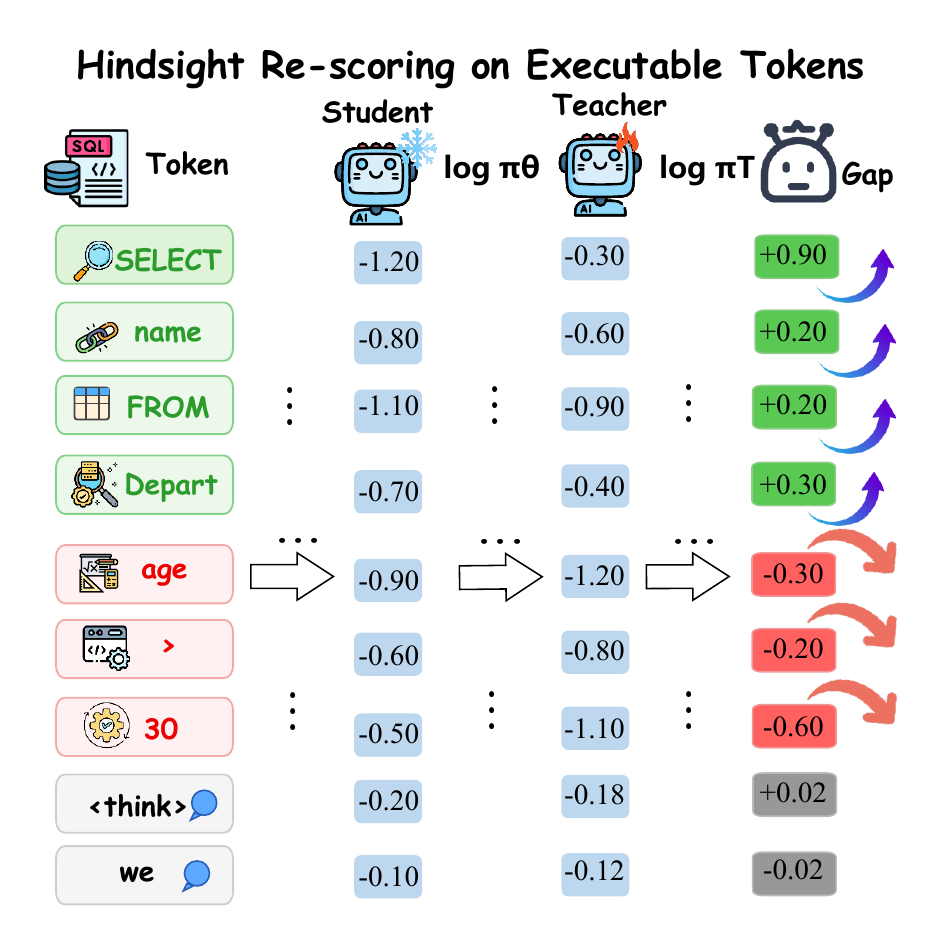}
  \caption{
    Re-scoring SQL tokens with execution hindsight and converts teacher and student likelihood gaps into masked token-level credit weights.
}
  \label{fig1:hindsight}
\end{figure}

\subsection{Selective Execution-Hindsight Teacher}

To obtain localized credit signals, we construct an execution-hindsight context for each SQL action displayed as Figure \ref{fig1:hindsight}. In our implementation, the hindsight context is derived from the immediate execution observation as
$c_t^{\mathrm{ex}}=H_{\mathrm{ex}}(o_t)
\label{eq:execution_context}
$,
where $c_t^{\mathrm{ex}}$ may contain the returned result, runtime or parser error, empty-result feedback, and other execution diagnostics.
This context is available only to the teacher during training and is not part of the student's decision-time input.

Let $\pi_T$ denote a teacher policy periodically synchronized with the student and treated as a stop-gradient model. For a student-generated token $y_{t,i}$, the teacher evaluates the same token with execution hindsight:
\begin{equation}
\log \pi_T
\left(
y_{t,i}
\mid
h_t,y_{t,<i},c_t^{\mathrm{ex}}
\right),
\label{eq:teacher_probability}
\end{equation}
whereas the student uses only the information available before taking the action:
\begin{equation}
\log \pi_\theta
\left(
y_{t,i}
\mid
h_t,y_{t,<i}
\right).
\label{eq:student_probability}
\end{equation}
The teacher does not generate an alternative trajectory. Instead, it re-scores the student's own on-policy actions using privileged
execution feedback.

Execution hindsight should affect only executable decisions. We therefore define an action mask
\begin{equation}
m_{t,i}\in\{0,1\},
\label{eq:action_mask}
\end{equation}
where $m_{t,i}=1$ when $y_{t,i}$ belongs to an executable SQL or tool action span, including SQL submission and revision actions, and
$m_{t,i}=0$ for reasoning, formatting, and other non-executable tokens.

\subsection{Hindsight-Guided Advantage Reweighting}

Let $A^n$ denote the group-relative advantage of trajectory $\tau^n$ defined in Eq.~\ref{eq:grpo_advantage}. Standard GRPO assigns $A^n$ to all tokens in the trajectory. To obtain token-level credit, we compute the teacher--student log-probability gap
\begin{equation}
\begin{aligned}
\Delta_{t,i}
={}&
\log \pi_T
\left(
y_{t,i}
\mid
h_t,y_{t,<i},c_t^{\mathrm{ex}}
\right)\\
&-
\log \pi_\theta
\left(
y_{t,i}
\mid
h_t,y_{t,<i}
\right).
\end{aligned}
\label{eq:teacher_student_gap}
\end{equation}
A positive $\Delta_{t,i}$ indicates that the sampled token becomes more plausible when the teacher observes execution hindsight, whereas a negative gap indicates that the token is less supported by the execution feedback.

We convert this gap into a bounded, sign-aware reweighting factor:
\begin{equation}
w_{t,i}
=
\operatorname{clip}
\left(
\exp\left(
\operatorname{sgn}(A^n)\,
\operatorname{sg}[\Delta_{t,i}]
\right),
w_{\min},
w_{\max}
\right),
\label{eq:hindsight_weight}
\end{equation}
where $\operatorname{sg}[\cdot]$ denotes the stop-gradient operator,and $w_{\min}=1-\epsilon_w$, $w_{\max}=1+\epsilon_w$. The sign of $A^n$ aligns the teacher evidence with the reward-derived update direction. For positive-advantage trajectories, hindsight- supported tokens receive stronger reinforcement. For negative-advantage trajectories, hindsight-disfavored tokens receive stronger suppression. When $A^n=0$, the reweighting factor reduces to one.

\begin{table*}[t]
\centering
\small
\resizebox{\textwidth}{!}{
\begin{tabular}{lcccccc}
\toprule
\textbf{Model} & \textbf{Size} & \textbf{Training Set} & \textbf{BIRD-Dev} & \textbf{Spider-Test} & \textbf{Spider2.0-SQLite} & \textbf{Spider-DK} \\
\midrule

\multicolumn{7}{c}{\textit{7B-scale models}} \\
CodeS & 7B & Spider & 57.17 & 85.40 & -- & 72.00 \\
SQL-R1 & 7B & SynSQL-2.5M + BIRD & 66.60 & 88.70 & -- & -- \\
ReEx-SQL & 7B & BIRD & 64.90 & 86.60 & -- & 79.80 \\
PaVeRL-SQL & 7B & BIRD & 69.30 & -- & \textbf{19.30} & -- \\
MARS-SQL & 7B & BIRD & \textbf{77.84} & 89.75 & -- & 78.13 \\
\textbf{SERL-SQL (Ours)} & \textbf{7B} & BIRD & \underline{75.55} & 89.24 & 14.47 & 80.39 \\

\midrule
\multicolumn{7}{c}{\textit{14B-scale models}} \\
Progress-SQL & 14B & BIRD & 67.50 & 88.10 & -- & 76.40 \\
Reasoning-SQL & 14B & BIRD & 72.29 & 81.43 & -- & 73.03 \\
\textbf{SERL-SQL (Ours)} & \textbf{14B} & BIRD & \underline{74.95} & \textbf{89.92} & \underline{25.64} & 81.68 \\

\midrule
\multicolumn{7}{c}{\textit{30B--32B-scale models and systems}} \\
FineStep & 30B & BIRD + Spider & 69.23 & 89.50 & -- & \underline{81.90} \\
OmniSQL & 32B & SynSQL-2.5M + Spider + BIRD & 64.50 & 87.60 & 11.90 & 76.10 \\
Arctic-Text2SQL-R1 & 32B & BIRD + Spider & 70.50 & 88.70 & 16.30 & \underline{80.60} \\
XiYanSQL & 32B & Unknown & 67.00 & -- & -- & -- \\
Agentar-Scale-SQL & 32B & BIRD & 74.90 & -- & -- & -- \\
\textbf{SERL-SQL (Ours)} & \textbf{32B} & BIRD & \underline{76.56} & \underline{89.78} & \textbf{26.78} & \textbf{82.21} \\
\bottomrule
\end{tabular}
}
\caption{Main results on BIRD-Dev, Spider-Test, Spider2.0-SQLite, and Spider-DK. All results are reported in execution accuracy (\%). The best and second-best results are highlighted in \textbf{bold} and \underline{underline}, respectively. ``--'' denotes unreported results.}
\label{tab:main_results}
\end{table*}

The selective action mask gives the effective token weight
\begin{equation}
\bar{w}_{t,i}
=
(1-m_{t,i})+m_{t,i}w_{t,i}.
\label{eq:masked_weight}
\end{equation}
Thus, non-action tokens retain the original GRPO weight, while executable SQL and tool-action tokens receive hindsight-guided
reweighting.

To reduce dependence on privileged teacher information, we introduce a training-dependent coefficient $\alpha_k\in[0,1]$ at optimization step $k$. The final token-level advantage is
\begin{equation}
\widetilde{A}_{t,i}
=
A^n
\left[
(1-\alpha_k)
+
\alpha_k\bar{w}_{t,i}
\right].
\label{eq:reweighted_advantage}
\end{equation}
The coefficient $\alpha_k$ is gradually decayed during training. Early updates use execution hindsight to improve credit assignment, whereas later updates increasingly return control to the reward-driven GRPO objective.

Using the clipped surrogate $\ell_{\mathrm{GRPO}}(\theta;B_{t,i})$ defined in Eq.~\ref{eq:grpo_surrogate}, the final objective is
\begin{equation}
\small
\begin{aligned}
\mathcal{L}_{\textsc{SERL-SQL}}(\theta)
={}&
-\mathbb{E}_{\tau}
\left[
\sum_{t,i}
\ell_{\mathrm{GRPO}}
\left(
\theta;\widetilde{A}_{t,i}
\right)
\right]\\
&+
\beta\,
\mathrm{KL}
\left(
\pi_\theta
\|
\pi_{\mathrm{ref}}
\right).
\end{aligned}
\label{eq:sql_serl_objective}
\end{equation}
The task reward therefore determines the direction of policy optimization, while execution hindsight adjusts only the locality and
magnitude of the update. No gradient is propagated through the teacher or the reweighting coefficient. At inference time, the teacher is removed and the trained student policy operates as a standalone multi-turn SQL agent.

\section{Experiments}
\subsection{Experimental Setup}
\noindent\textbf{Setting.}
We evaluate our method on the BIRD~\cite{li2023bird}, Spider~\cite{yu2018spider}, Spider-DK~\cite{gan2021exploringunderexploredlimitationscrossdomainspiderdk}, and Spider 2.0-lite~\cite{lei2025spider} sqlite benchmark, a particularly challenging cross-domain dataset. And we deployed system contains three specialized 7B, 14B, and 32B model.

\noindent\textbf{Baselines.}
We compared several top-ranking baseline methods from the overall leaderboard. The former consists of fifteen baselines, including CodeS~\cite{10.1145/3654930}, SQL-R1~\cite{ma2025sqlr1trainingnaturallanguage}, ReEx-SQL~\cite{dai-etal-2026-reex}, PaVeRL-SQL~\cite{hao2025paverlsqltexttosqlpartialmatchrewards}, XiYan-SQL~\cite{gao2024xiyansql}, among others. The latter comprises eight leading methods, such as Agentar-Scale-SQL\cite{wang2025agentarscalesql}, Progress-SQL~\cite{zhang2026progresssqlimprovingreinforcementlearning}, Reasoning-SQL~\cite{pourreza2025reasoningsql}, FineStep~\cite{dai2026stepcountssteplevelcredit}, OmniSQL~\cite{10.14778/3749646.3749723}, Arctic-Text2SQL-R1-32B~\cite{yao2026arctictext2sqlr1simplerewardsstrong}, MARS-SQL~\cite{yang2025marssql}.

\noindent\textbf{Evaluation Metrics.}
Following prior work~\cite{pourreza2024chasesql}, we use Execution Accuracy (EX), the official metric for the respective leaderboard.

\noindent\textbf{Experiment Implementations.}
All models are based on Qwen2.5-Coder-Instruct~\cite{hui2024qwen25codertechnicalreport} including 7B,14B,32B, implemented in PyTorch and trained on 16 NVIDIA H100 GPUs or 8 NVIDIA H200 GPUs. The Grounding Schema Agent utilize qwen3-Embedding-0.6B and Qwen2.5-Coder-7B-Instruct for schema linking. Qwen2.5-Coder models are used for training and inference and detailed hyperparameters or prompts are provided in Appendix.

\begin{table}[t]
\centering
\small
\setlength{\tabcolsep}{0.7mm}
\resizebox{\columnwidth}{!}{
\begin{tabular}{lcccc}
\toprule
\textbf{Method} &
\textbf{BIRD-Dev} &
\textbf{Spider-Test} &
\textbf{Spider-DK}  \\
\midrule
w/o Schema Grounding       & 66.42 & 84.39 & 71.95  \\
w/o Execution Hindsight    & 71.94 & 87.37 & 78.65  \\
w/o Weight Clipping  & 73.14 & 88.66 & 79.89  \\
w/o Selective Action Mask & 73.43 & 88.87 & 80.14  \\
w/o Hindsight Decay        & 73.61 & 89.01 & 80.12  \\
w/o Sign-Aware Reweighting    & 73.92 & 89.22 & 80.64  \\
\midrule
\textbf{SERL-SQL(Full)}    & \textbf{74.95} & \textbf{89.92} & \textbf{81.68}  \\
\bottomrule
\end{tabular}
}
\caption{Ablation results of SERL-SQL on benchmarks.}
\label{tab:ablation}
\end{table}

\subsection{Experimental Results}
As shown in Table~\ref{tab:main_results}, SERL-SQL achieves competitive performance across both in-domain and cross-domain benchmarks. On BIRD-Dev, SERL-SQL obtains 75.55\% with the 7B model and 74.95\% with the 14B model, outperforming several strong RL-based and reasoning-enhanced baselines such as Reasoning-SQL, Arctic-Text2SQL-R1, and Agentar-Scale-SQL, while remaining close to MARS-SQL. On Spider-Test, SERL-SQL on 14B model achieves the best result of 89.92\%, and the 7B variant also reaches a competitive accuracy of 89.24\%. Notably, SERL-SQL is trained only on BIRD, whereas several competing methods rely on additional Spider, synthetic, or constructed SQL data. Scaling to 32B further strengthens the overall performance, reaching 76.56\% on BIRD-Dev, 26.78\% on Spider2.0-SQLite, and 82.21\% on Spider-DK, showing that execution-hindsight-guided credit assignment remains effective at larger model scales and challenging cross-domain settings. More experimental details are provided in the Appendix.

\subsection{Further Study}
\noindent\textbf{Analysis by Component.}
Table~\ref{tab:ablation} verifies the contribution of each component in SERL-SQL. Removing schema grounding leads to the largest drop, showing that effective credit assignment depends on a compact and semantically aligned schema context; noisy schema information can degrade both SQL generation and hindsight interpretation. Among the training-side modules, execution hindsight is the most critical, confirming that execution feedback provides localized evidence beyond sparse trajectory-level rewards. The remaining ablations show that this feedback must be carefully controlled: action masking prevents credit from diffusing to non-executable tokens, sign-aware reweighting preserves the reward-determined update direction, and clipping with decay limits noisy or privileged teacher signals. Overall, SERL-SQL gains come from converting execution feedback into bounded, localized, and reward-aligned token-level credit, rather than simply adding an auxiliary teacher signal.

\begin{figure}[t]
  \centering
  \includegraphics[width=\linewidth]{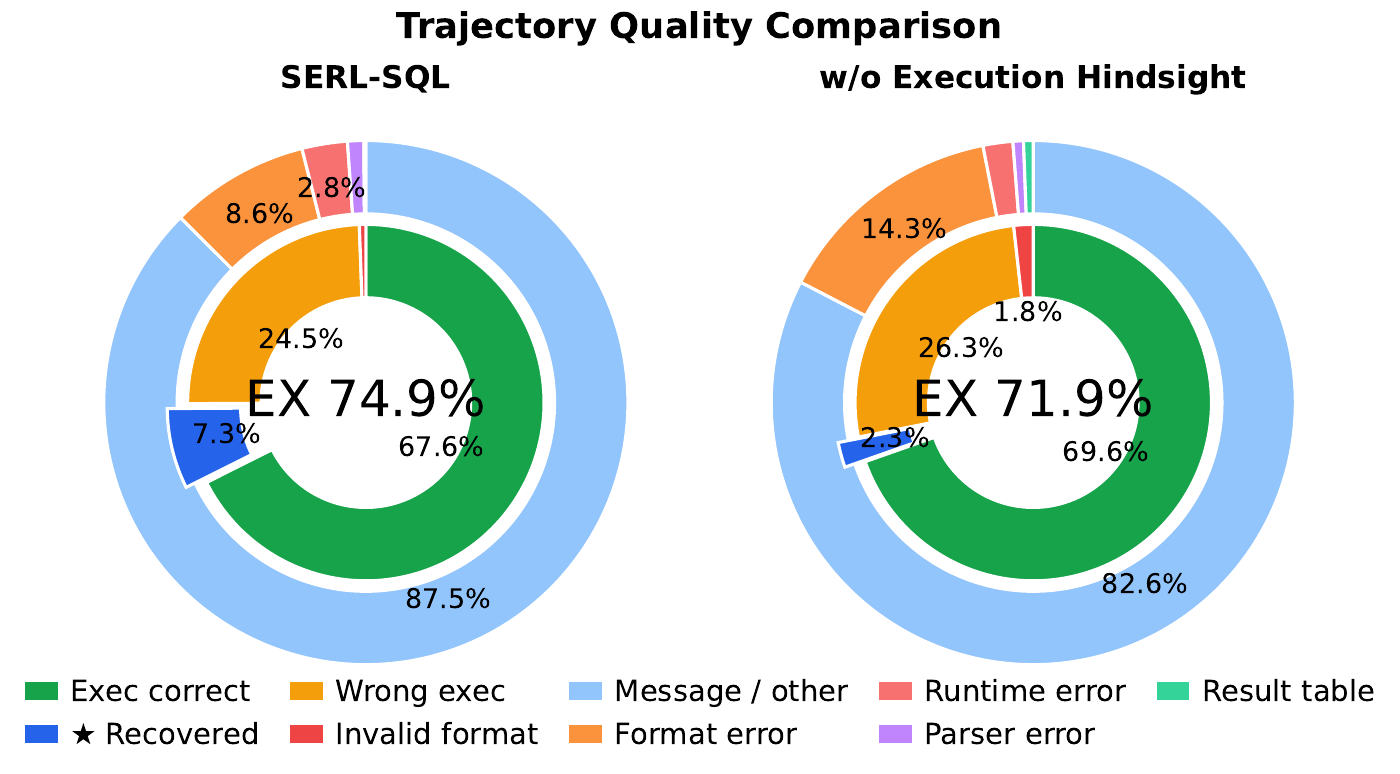}
  \caption{Trajectory quality on BIRD-dev.
    \textbf{Inner ring:} final outcome under selection for direct exec-correct, recovered, wrong exec, invalid format, \textbf{Outer ring:} observation feedback types aggregated over all interaction steps.}
  \label{fig:trajectory_quality}
\end{figure}

\begin{figure*}[tb]
  \centering
  \includegraphics[width=\linewidth]{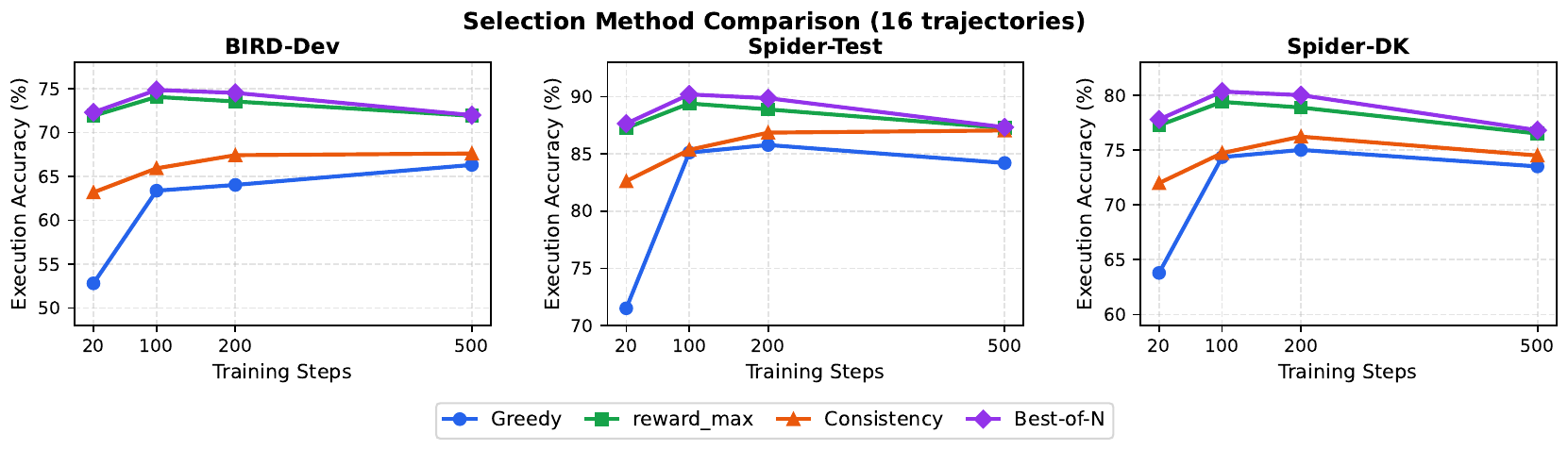}
  \caption{
    Selection method comparison with 16 sampled trajectories. \textbf{Best-of-N} denotes oracle selection, \textbf{reward\_max} selects the highest-reward trajectory, \textbf{greedy} uses a single rollout, and \textbf{consistency} selects the majority answer.
}
  \label{fig:selection_compare}
\end{figure*}

\noindent\textbf{Analysis by Selection Methods.}
Figure~\ref{fig:selection_compare} compares different selection strategies under 16 sampled trajectories. Across BIRD-Dev, Spider-Test, and Spider-DK, \textit{Best-of-N} consistently provides the upper bound, while \textit{reward\_max} closely tracks this oracle performance and substantially outperforms greedy decoding, especially in the early training stage. This suggests that SERL-SQL does not only improve the single sampled trajectory, but also increases the probability of generating high-quality candidates that can be identified by execution-based reward selection. In contrast, consistency-based selection is generally weaker than reward\_max, indicating that majority agreement is less reliable for Text-to-SQL when multiple trajectories may converge to the same executable but semantically incorrect query. The performance peak around intermediate training steps further suggests a trade-off between policy improvement and candidate diversity: longer training improves the base policy, but may reduce the diversity needed for effective multi-trajectory selection. More experimental details are provided in the Appendix.

\begin{table}[t]
\centering
\setlength{\tabcolsep}{0.5mm}
\resizebox{\columnwidth}{!}{
\begin{tabular}{lccc}
\toprule
\textbf{Hindsight Source(\%)} &
\textbf{BIRD-Dev} &
\textbf{Spider-Test} &
\textbf{Spider-DK} \\
\midrule
None                  & 73.58 & 87.98 & 79.74 \\
Error Type Only       & 74.16 & 88.56 & 80.32 \\
Error Message Only    & 74.87 & 89.27 & 81.03 \\
Result Table Only     & 74.12 & 88.52 & 80.28 \\
Immediate Observation & \textbf{75.52} & \textbf{89.92} & \textbf{81.68} \\
Full Observation      & 74.28 & 88.68 & 80.44 \\
\bottomrule
\end{tabular}
}
\caption{Effect of different execution-hindsight sources.}
\label{tab:hindsight_source}
\end{table}

% \begin{figure}[ht]
%   \centering
%   \includegraphics[width=\linewidth]{AnonymousSubmission/LaTeX/hindsight_decay_plot.pdf}
%   \caption{
%     Hindsight teacher-signal decay analysis on BIRD-Dev. A moderate decay horizon $K_{\mathrm{decay}}=100$ achieves the best reward\_max EX, outperforming both no-hindsight and non-decayed teacher settings. This indicates that execution hindsight is most useful as transient credit guidance rather than persistent privileged supervision.
%     }
%   \label{fig:hindsight_decay}
% \end{figure}

\noindent\textbf{Analysis by Trajectory Quality.}
Figure~\ref{fig:trajectory_quality} shows that execution hindsight improves both final accuracy and trajectory reliability. SERL-SQL achieves a higher EX of 74.9\% compared with 71.9\% without execution hindsight, mainly by increasing the proportion of recovered trajectories and reducing format errors. Although the no-hindsight variant has a slightly higher ratio of directly execution-correct cases, it produces fewer successful recoveries and more malformed interactions, suggesting that it is less effective at using feedback during multi-turn revision. In contrast, SERL-SQL better exploits execution observations to revise incorrect intermediate SQL decisions, while maintaining more stable interaction formats. These results indicate that execution-hindsight credit assignment improves not only final selection accuracy, but also the agent's ability to produce recoverable and executable multi-turn trajectories.

\subsection{Case Study}
\noindent\textbf{Analysis by Hindsight Source.}
Table~\ref{tab:hindsight_source} shows that execution hindsight is most effective when it is causally aligned with the current SQL action. All hindsight variants outperform the no-hindsight setting, confirming that execution feedback provides useful local evidence beyond sparse trajectory-level rewards. Immediate Observation achieves the best performance because it directly reflects the consequence of the current action, preserving a clear action--feedback correspondence. In contrast, Full Observation brings no further gain, suggesting that longer feedback may introduce redundant or weakly related signals that blur token-level attribution. These results support SERL-SQL's design principle: hindsight should be localized, action-relevant, and diagnostic rather than simply richer. More experimental details are provided in the Appendix.

\begin{figure}[bt]
  \centering
  \includegraphics[width=\linewidth]{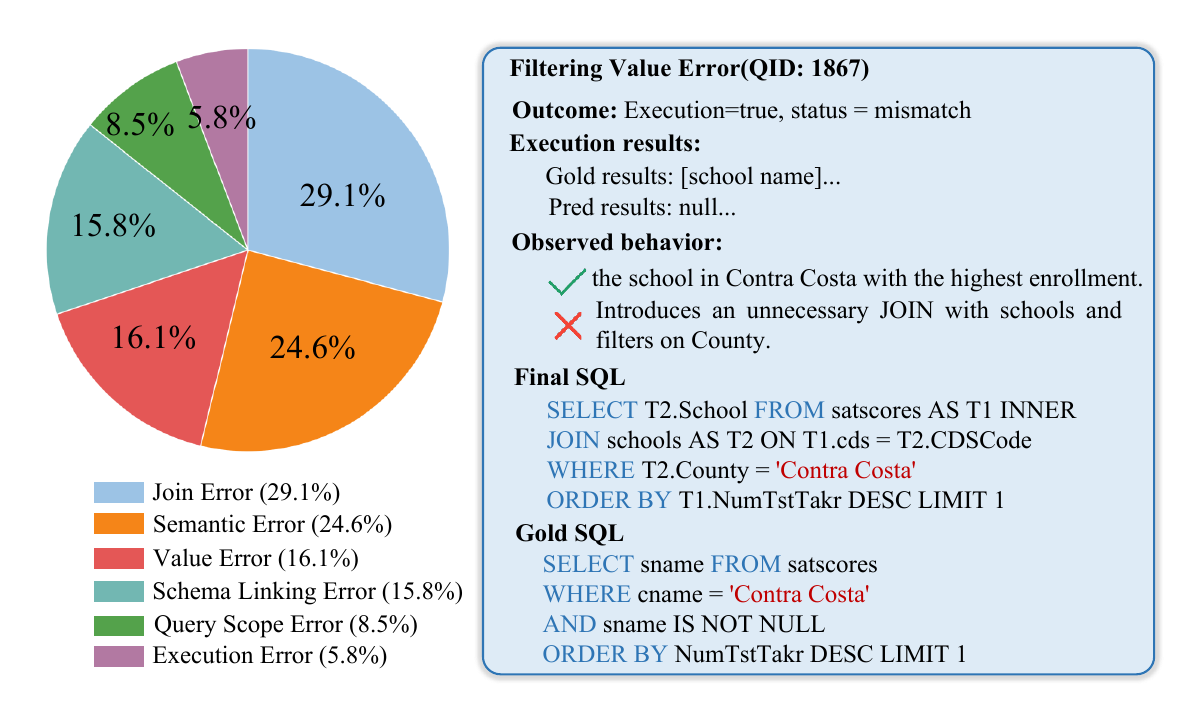}
  \caption{
    Error distribution and representative failure Case. 
    % The left pie chart reports the major error categories, while the right panel illustrates a value-filtering error where the model generates an executable SQL query but fails to align the predicted filtering condition with the gold SQL.
}
  \label{fig:error}
\end{figure}

\noindent\textbf{Analysis by Error Cases.}
Figure~\ref{fig:error} shows that SERL-SQL's remaining errors are mainly caused by join construction, semantic misunderstanding, value filtering, and schema linking. Join errors account for the largest proportion, indicating that multi-table reasoning remains a key bottleneck even when execution feedback is available. The case study further shows a typical value-filtering failure: although the generated SQL is executable and structurally close to the gold query, it selects a different filtering condition, leading to an incorrect result. This suggests that future improvements should strengthen fine-grained value grounding and condition alignment, rather than only improving SQL executability. More experimental details are provided in the Appendix.

\section{Related Work}
\subsection{On-Policy Distillation and Self-Distillation}
On-policy distillation~\cite{agarwal2024onpolicy} improves language model policies by sampling trajectories from the current policy and using a teacher to provide dense supervision on the states actually visited by the student, thereby reducing the distribution mismatch of off-policy
distillation~\cite{tan2026selfsupervisedonpolicydistillationreasoning,li2026rethinkingonpolicydistillationlarge}. On-policy self-distillation~\cite{zhao2026selfdistilledreasoner} further removes the need for an external teacher by letting the same model play asymmetric roles, where the teacher branch conditions on privileged information while the student branch observes only deployable inputs~\cite{
zhao2026selfdistilledreasoner,lu2026sdar}. However, recent studies show that naively applying teacher guidance can be unstable under long-horizon rollout drift, teacher--student mismatch, or
privileged-information bias~\cite{
fu2026revisitingonpolicydistillationempirical,
yu2026dopddualonpolicydistillation,
xing2026trustregiononpolicydistillation}. Recent multi-turn agent methods constrain OPD through RL schedules, turn-level guidance, or selective hindsight placement, applying teacher signals only to reward-aligned action regions~\cite{li2026serl,lu2026sdar,li2026onpolicydistillationcurriculumturnlevel,tan2026selfsupervisedonpolicydistillationreasoning}. Our work extends this principle to the structured Text-to-SQL setting, where database execution feedback is converted into bounded, reward-aligned credit signals over SQL-relevant and tool-action tokens, rather than used as a full-response imitation target.

\subsection{Agentic Text-to-SQL Systems}

Recent Text-to-SQL systems improve performance through specialized training, data scaling, reinforcement learning, and database interaction. CodeS~\cite{10.1145/3654930} and OmniSQL~\cite{10.14778/3749646.3749723} strengthen SQL generation through domain-specific pre-training and large-scale synthetic data, while XiYan-SQL~\cite{gao2024xiyansql} and Agentar-Scale-SQL~\cite{wang2025agentarscalesql} improve candidate coverage through ensemble generation, iterative refinement, and test-time scaling. SQL-R1~\cite{ma2025sqlr1trainingnaturallanguage} and Arctic-Text2SQL-R1~\cite{yao2026arctictext2sqlr1simplerewardsstrong} optimize models with execution-based rewards, whereas Reasoning-SQL~\cite{pourreza2025reasoningsql} and PaVeRL-SQL~\cite{hao2025paverlsqltexttosqlpartialmatchrewards} introduce partial rewards to reduce reward sparsity. Some works~\cite{yang2025marssql,dai-etal-2026-reex,xu2025mtirsqlmultiturntoolintegratedreasoning} incorporate execution feedback into iterative revision, while Progress-SQL~\cite{zhang2026progresssqlimprovingreinforcementlearning} and other token-level works~\cite{dai2026stepcountssteplevelcredit,jian2026trustsqltoolintegratedmultiturnreinforcement} provide progressive or step-level supervision. Despite these advances, execution feedback is still largely assigned at the trajectory or step level, leaving precise credit assignment for executable SQL decisions underexplored.

\section{Conclusion}

We present SERL-SQL, which uses database feedback as training-time hindsight to selectively reweight SQL/tool-action tokens under reward-aligned RL. Experiments demonstrate that this localized execution-aware credit assignment improves Text-to-SQL generation and generalization.

\bibliography{aaai2027}


\appendix

\section{Implementation Details}

\subsection{Analysis by other models.}
Table~\ref{tab:selection_model_compare} compares trajectory selection strategies across different backbone models on BIRD. Overall, \textit{reward\_max} consistently outperforms consistency-based selection across all models, showing that execution-grounded rewards provide a more reliable correctness signal than majority agreement. More importantly, \textit{reward\_max} closely approaches the oracle \textit{Best-of-N} performance, especially for stronger SQL-specialized models such as XiYanSQL-QwenCoder-7B/14B and Qwen2.5-Coder-14B-Instruct, where the gap is within a small margin. This indicates that SERL-SQL not only increases the chance of generating correct candidates, but also makes those candidates identifiable by lightweight execution-based rewards. In contrast, consistency selection remains weaker, suggesting that multiple trajectories may agree on the same executable but semantically incorrect SQL. These results support the use of reward-based trajectory selection as an effective and low-cost approximation to oracle Best-of-N selection.

\begin{table}[t]
\centering
\small
\setlength{\tabcolsep}{0.5mm}
\begin{tabular}{lccc}
\toprule
\textbf{Model} &
\textbf{reward\_max} &
\textbf{Consistency} &
\textbf{Best-of-N} \\
\midrule
Qwen2.5-Coder-3B-Instruct   & 62.06 & 52.02 & 63.62 \\
Qwen2.5-7B-Instruct        & 68.88 & 60.39 & 73.16 \\
OmniSQL-7B                & 70.90 & 66.78 & 71.03 \\
OmniSQL-14B                & 71.25 & 67.98 & 71.87 \\
XiYanSQL-QwenCoder-7B       & 74.75 & 65.91 & 74.89 \\
XiYanSQL-QwenCoder-14B                & \textbf{75.53} & 67.41 & \textbf{75.78} \\
Qwen2.5-Coder-14B-Instruct        & 74.90 & 65.25 & 74.97 \\
Qwen3-8B        & 70.17 & 68.03 & 71.78 \\
Qwen3-14B        & 71.36 & \textbf{68.28} & 72.85 \\
\bottomrule
\end{tabular}
\caption{Comparison of trajectory selection strategies under different model settings on \textbf{Bird}. All results are reported in execution accuracy (\%).}
\label{tab:selection_model_compare}
\end{table}

\subsection{Analysis by trajectories.}
Figure~\ref{fig:trajectory_count_compare} studies the effect of the number of sampled trajectories at 100 training steps. Across BIRD-Dev, Spider-Test, and Spider-DK, \textit{reward\_max} consistently tracks the oracle \textit{Best-of-N} curve and clearly outperforms greedy and consistency-based selection, confirming that execution rewards provide an effective signal for identifying high-quality SQL candidates. The best performance is generally achieved around 16 trajectories, after which increasing the sampling budget brings little or even negative gain. This suggests that SERL-SQL benefits from moderate candidate diversity, but excessive sampling may introduce more noisy or semantically misleading executable queries, making selection harder. In contrast, consistency remains weaker because majority agreement can favor common but incorrect SQL patterns. Overall, the results show that lightweight reward-based selection provides a strong accuracy--cost trade-off with only a moderate number of trajectories.

\subsection{Analysis by different turns.}
Figure~\ref{fig:turn_count_compare} analyzes the effect of the maximum interaction turns at 100 training steps. Across all three benchmarks, performance generally improves from 1 to 5 turns, indicating that moderate multi-turn interaction helps the agent use execution feedback to revise intermediate SQL decisions. The best results appear around $T=5$, where \textit{reward\_max} remains close to the oracle \textit{Best-of-N} upper bound and consistently outperforms greedy and consistency-based selection. However, increasing the turn budget to $T=10$ does not further improve performance and even causes slight degradation, suggesting that overly long interactions may introduce redundant revisions, accumulated formatting errors, or noisy executable candidates. These results show that SERL-SQL benefits from multi-turn execution feedback, but only up to a moderate horizon where feedback remains informative and controllable.

\begin{figure}[ht]
  \centering
  \includegraphics[width=\linewidth]{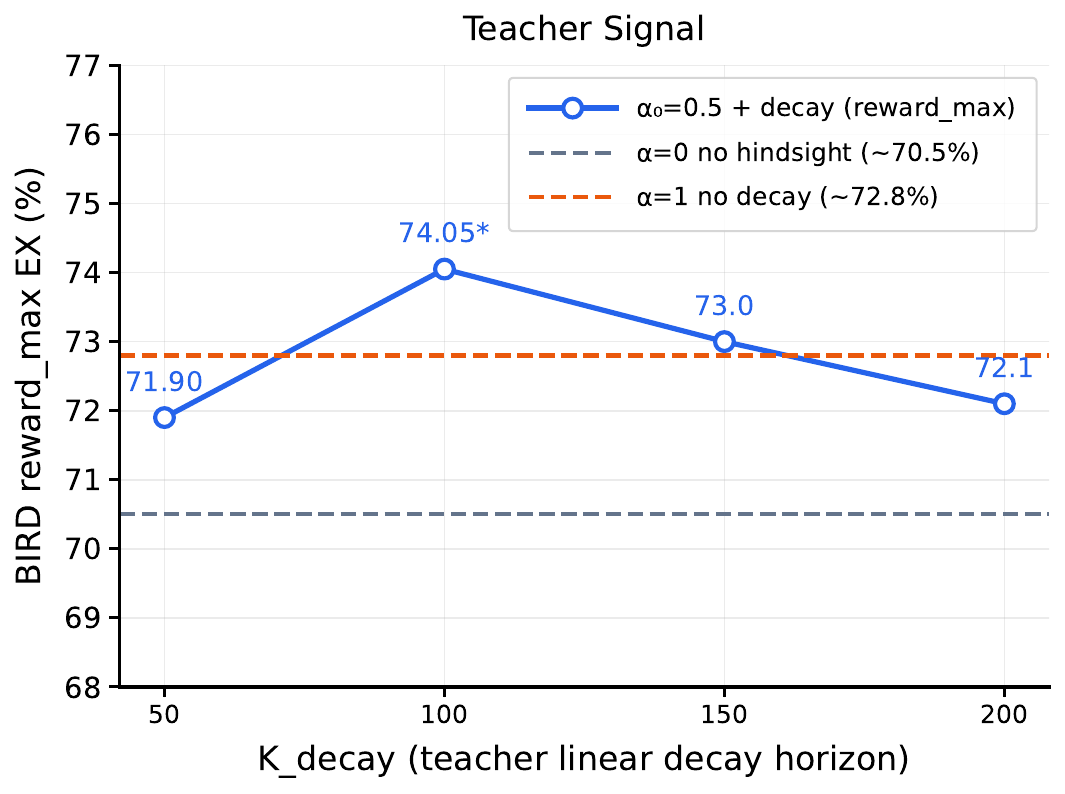}
  \caption{
    Hindsight teacher-signal decay analysis on BIRD-Dev. A moderate decay horizon $K_{\mathrm{decay}}=100$ achieves the best reward\_max EX, outperforming both no-hindsight and non-decayed teacher settings. This indicates that execution hindsight is most useful as transient credit guidance rather than persistent privileged supervision.
    }
  \label{fig:hindsight_decay}
\end{figure}

\subsection{Analysis by Hindsight Decay.}
The decay horizon controls the balance between dense credit assignment and privileged-teacher bias. As shown in Figure~\ref{fig:hindsight_decay}, removing hindsight guidance yields only 70.5\% BIRD reward\_max EX, indicating that trajectory-level execution rewards are too coarse for multi-turn SQL credit assignment. A non-decayed teacher signal improves performance to 72.8\%, but still falls behind the decayed variants, suggesting that execution hindsight is useful but should not remain a persistent optimization target. The best result is achieved at $K_{\mathrm{decay}}=100$, where SERL-SQL reaches 74.05\%. This moderate schedule uses teacher-induced likelihood gaps to shape early updates when exploration is weak, and then gradually returns control to the reward-driven GRPO objective as the policy becomes more reliable. Shorter decay removes hindsight before it is fully exploited, whereas longer decay may retain privileged or noisy teacher signals for too long. These results support using execution hindsight as transient credit guidance rather than permanent distillation supervision.

\subsection{Analysis by Error Cases.}
\paragraph{Case Analysis.}
We further analyze representative failure cases to better understand the
remaining limitations of SERL-SQL.

\begin{itemize}
    \item \textbf{Execution error.}
    Figure~\ref{fig:execution_case} shows a low-level SQL realization
    failure. The model captures the intended join structure, but uses
    \texttt{order}, a reserved SQL keyword, as an unquoted table name,
    making the query non-executable. This suggests that some failures
    are caused by fragile SQL surface realization rather than incorrect
    task understanding.

    \item \textbf{Join error.}
    Figure~\ref{fig:join_case} illustrates a join-path error. Although
    the model identifies the target entity, it introduces an unnecessary
    join and applies the filtering condition to the wrong table. This
    indicates that multi-table path construction remains challenging
    when similar semantic evidence appears across related tables.

    \item \textbf{Query-scope error.}
    Figure~\ref{fig:query_scope_case} shows a scope-control failure. The
    model retrieves the correct attribute but omits a required non-null
    constraint, producing an overly broad result set. This case highlights
    that Text-to-SQL correctness often depends on subtle constraints
    beyond table and column selection.

    \item \textbf{Schema-linking error.}
    Figure~\ref{fig:schema_case} presents an incorrect field-grounding
    case. The model captures the user and temporal condition, but links
    to a nonexistent column, leading to a runtime error. This suggests
    that precise schema grounding is still critical, especially for
    databases with semantically similar fields.

    \item \textbf{Semantic error.}
    Figure~\ref{fig:semantic_case} shows that an executable SQL query can
    still be semantically incorrect. The model applies the correct filter
    but projects an extra column, causing the output schema to mismatch
    the question requirement.

    \item \textbf{Value error.}
    Figure~\ref{fig:value_case} highlights a value-normalization failure.
    The model uses a natural-language value directly instead of mapping
    it to the database-specific code, and also mishandles the temporal
    range. This suggests that future work should strengthen value
    grounding, database-specific normalization, and fine-grained
    condition alignment.
\end{itemize}

Overall, these cases show that the remaining errors mainly arise from
localized executable decisions, including SQL realization, join-path
selection, scope control, schema grounding, projection choice, and value
normalization. This supports the motivation of SERL-SQL: execution
feedback is valuable not only as an outcome reward, but also as a signal
for identifying which SQL actions require stronger or weaker updates.

\begin{figure*}[tb]
  \centering
  \includegraphics[width=\linewidth]{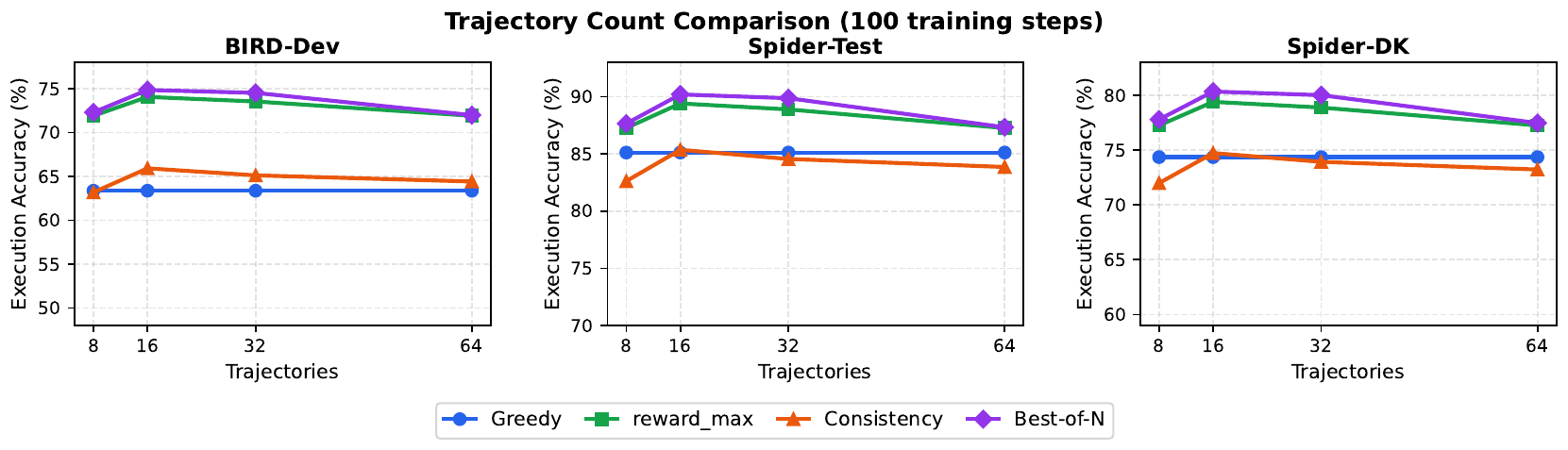}
  \caption{
    Selection method comparison with different sampled trajectories. \textbf{Best-of-N} denotes oracle selection, \textbf{reward\_max} selects the highest-reward trajectory, \textbf{greedy} uses a single rollout, and \textbf{consistency} selects the majority answer.
}
  \label{fig:trajectory_count_compare}
\end{figure*}

\begin{figure*}[tb]
  \centering
  \includegraphics[width=\linewidth]{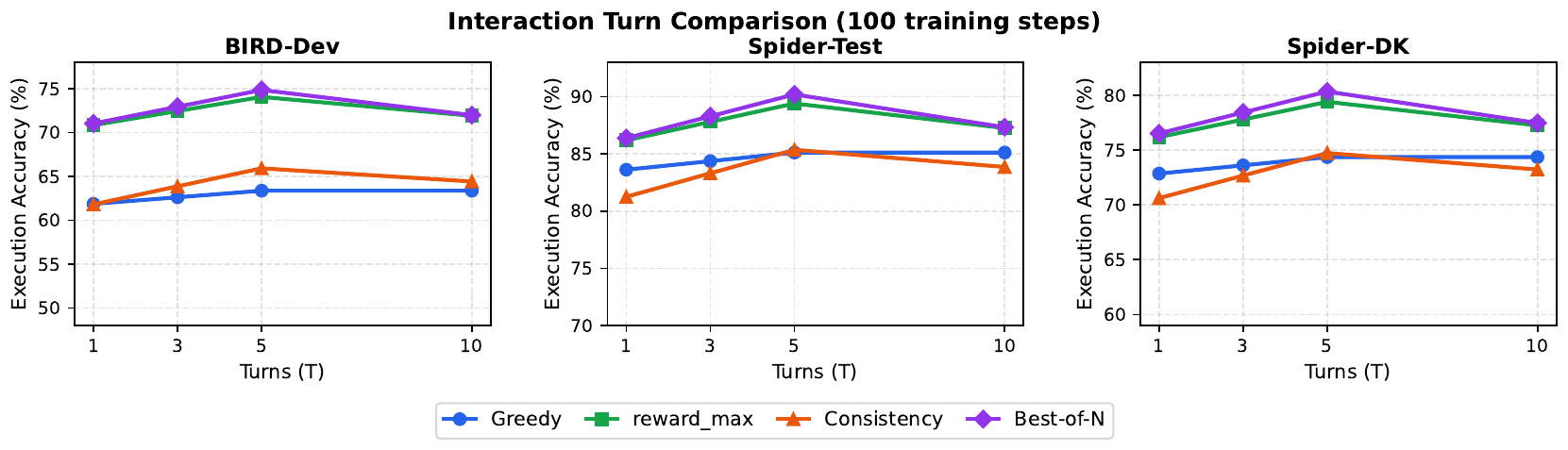}
  \caption{
    Selection method comparison with different agent turns. \textbf{Best-of-N} denotes oracle selection, \textbf{reward\_max} selects the highest-reward trajectory, \textbf{greedy} uses a single rollout, and \textbf{consistency} selects the majority answer.
}
  \label{fig:turn_count_compare}
\end{figure*}

\section{Experimental Details}
\subsection{Details of Public Baselines}
\label{sec:pubmethod}

We compare SERL-SQL with a diverse set of publicly available Text-to-SQL
methods covering SQL-specialized pre-training, large-scale data synthesis,
reinforcement learning, execution-aware interaction, fine-grained credit
assignment, multi-agent reasoning, and test-time scaling. We briefly summarize
the main characteristics of these baselines below.

\begin{itemize}

    \item \textbf{CodeS:}
    CodeS~\cite{li2024codes} is a family of open-source language models specialized for Text-to-SQL.
    It improves SQL generation through incremental pre-training on an SQL-centric
    corpus, together with schema-aware prompting and data augmentation for
    schema linking and domain adaptation.

    \item \textbf{SQL-R1:}
    SQL-R1~\cite{ma2025sqlr1trainingnaturallanguage} trains reasoning-oriented Text-to-SQL models with reinforcement
    learning and SQL-specific reward signals. It combines reward-driven policy
    optimization with synthetic Text-to-SQL data to improve reasoning over
    complex queries such as multi-table joins and nested SQL.

    \item \textbf{ReEx-SQL:}
    ReEx-SQL~\cite{dai-etal-2026-reex} introduces execution-aware reinforcement learning by interleaving
    SQL generation with database execution during reasoning. Execution feedback
    is incorporated into stepwise rollouts to support online query revision,
    while an exploration-oriented reward and tree-based decoding encourage
    effective database interaction.

    \item \textbf{PaVeRL-SQL:}
    PaVeRL-SQL~\cite{hao2025paverlsqltexttosqlpartialmatchrewards} combines partial-match rewards with verbal reinforcement learning.
    It provides intermediate feedback beyond binary execution correctness and
    supports both verbal self-improvement and reinforcement-learning-based
    reasoning for complex Text-to-SQL tasks.

    \item \textbf{MARS-SQL:}
    MARS-SQL~\cite{yang2025marssql} is a multi-agent Text-to-SQL framework consisting of specialized
    grounding, generation, and validation agents. Its generation agent is
    optimized with multi-turn reinforcement learning and interacts with the
    database through a Think--Act--Observe loop, enabling iterative SQL
    refinement from execution feedback.

    \item \textbf{Progress-SQL:}
    Progress-SQL~\cite{zhang2026progresssqlimprovingreinforcementlearning} formulates Text-to-SQL as a multi-turn reinforcement learning
    problem with progressive rewards. It evaluates the improvement between
    intermediate and refined SQL queries using structural and execution-aware
    signals, thereby providing denser supervision for iterative correction.

    \item \textbf{Reasoning-SQL:}
    Reasoning-SQL~\cite{pourreza2025reasoningsql}  applies GRPO with SQL-tailored partial rewards, including
    schema-linking, syntax, semantic-feedback, and query-similarity signals.
    These intermediate rewards alleviate the sparsity of pure execution
    supervision and encourage the model to develop stronger SQL reasoning
    capabilities.

    \item \textbf{FineStep:}
    FineStep~\cite{dai2026stepcountssteplevelcredit} addresses credit assignment in tool-integrated Text-to-SQL by
    assigning supervision at the reasoning-step level. It combines independent
    process rewards with step-level advantage estimation, allowing policy
    updates to distinguish useful intermediate interactions from redundant or
    erroneous steps.

    \item \textbf{OmniSQL:}
    OmniSQL~\cite{10.14778/3749646.3749723} is an open-source Text-to-SQL model family built upon a scalable
    synthetic-data generation framework. It introduces SynSQL-2.5M, a
    large-scale corpus spanning diverse databases and SQL structures, and
    combines synthetic and human-annotated data to improve cross-domain SQL
    generation.

    \item \textbf{Arctic-Text2SQL-R1:}
    Arctic-Text2SQL-R1~\cite{yao2026arctictext2sqlr1simplerewardsstrong} is a reinforcement-learning framework that optimizes
    Text-to-SQL reasoning using a lightweight reward based primarily on
    execution correctness. Rather than relying on complex intermediate reward
    shaping, it combines execution-based RL with curated training data and
    supervised initialization.

    \item \textbf{XiYan-SQL:}
    XiYan-SQL~\cite{gao2024xiyansql} is a multi-generator ensemble framework that combines schema
    filtering, diverse SQL candidate generation, and learned candidate
    selection. Multiple generators are used to improve candidate diversity,
    followed by a selector that identifies the final SQL prediction.

    \item \textbf{Agentar-Scale-SQL:}
    Agentar-Scale-SQL~\cite{wang2025agentarscalesql} improves Text-to-SQL through orchestrated test-time
    scaling. It jointly exploits RL-enhanced intrinsic reasoning, sequential
    iterative refinement, and parallel candidate generation and selection to
    scale inference-time computation from complementary perspectives.

\end{itemize}

\paragraph{Hyperparameters Details.}
We provide the detailed implementation settings of SERL-SQL in Tables~1--4. Table~1 summarizes the main training hyperparameters, including the training steps, rollout group size, optimizer, learning rate, PPO settings, and memory-saving strategies. Table~2 reports the SERL-SQL-specific configurations for execution-hindsight reweighting, such as the teacher synchronization interval, hindsight source, action mask, weight clipping range, and hindsight decay. Table~3 lists the rollout, decoding, and trajectory selection settings used during training and evaluation, including sampling temperature, top-$p$, maximum sequence lengths, and selection strategies. Finally, Table~4 describes the reward and verification protocol, including format constraints, execution rewards, SQL execution timeout, and result comparison rules.

\begin{table}[htbp]
\centering
\small
\setlength{\tabcolsep}{4pt}
\begin{tabular}{ll}
\toprule
\textbf{Hyperparameter} & \textbf{Value} \\
\midrule
Training steps & 100 for platform 14B run; \\
Rollout group size $G$ & 8 or 16 or 32 \\
Sampled trajectories per update & batch size $\times G$ \\
Maximum interaction turns & 5 by default \\
Learning rate & $1\times 10^{-6}$ \\
Optimizer & AdamW \\
Weight decay & 0.01 \\
PPO mini-batch size & Same as train batch size \\
PPO micro-batch size & 8 \\
Log-prob micro-batch size & 8 \\
Policy clip range & $[1-0.2, 1+0.2]$ \\
Gradient clipping & 0.5 \\
Entropy coefficient & 0 \\
KL loss & Disabled in training script \\
KL coefficient if enabled & 0.001 \\
KL loss type & low\_var\_kl \\
Gradient checkpointing & Enabled \\
Parameter offloading & Enabled \\
Optimizer offloading & Enabled \\
\bottomrule
\end{tabular}
\caption{Training hyperparameters used by SERL-SQL.}
\label{tab:appendix_training_hparams}
\end{table}

\begin{table}[htbp]
\centering
\small
\setlength{\tabcolsep}{4pt}
\begin{tabular}{ll}
\toprule
\textbf{Hyperparameter} & \textbf{Value} \\
\midrule
Policy loss mode & serl\_action\_mask \\
Teacher type & Synchronized training-only teacher \\
Teacher synchronization interval & 10 steps \\
Hindsight source & Immediate execution feedback \\
Sampling mode & immediate\_feedback \\
Initial hindsight coefficient $\alpha_0$ & 0.5 \\
Hindsight decay horizon & Same as total training steps \\
Weight clipping range & $[0.8, 1.2]$ \\
Sign-aware reweighting & Enabled \\
Action mask mode & SQL/action tokens only \\
Successful trajectory context & Included \\
Immediate feedback context & Included \\
Success reward threshold & 1.0 \\
Self-success reprompting & Disabled \\
Thinking trace in demonstration & Removed \\
Non-action tokens & Keep original GRPO weight \\
Teacher gradient & Stop-gradient \\
\bottomrule
\end{tabular}
\caption{SERL-SQL-specific hyperparameters for execution-hindsight reweighting.}
\label{tab:appendix_serl_hparams}
\end{table}

\begin{table}[htbp]
\centering
\small
\setlength{\tabcolsep}{4pt}
\begin{tabular}{ll}
\toprule
\textbf{Item} & \textbf{Value} \\
\midrule
Training rollout temperature & 0.6 \\
Training rollout top-$p$ & 0.95 \\
Training rollout group size & 8 \\
Evaluation sampled trajectories & 16 \\
Evaluation temperature & 0.8 \\
Evaluation top-$p$ & 0.7 \\
Evaluation sampling & Enabled \\
Greedy evaluation & Temperature 0 \\
Maximum prompt length during training & 4096 \\
Maximum prompt length during evaluation & 8192 \\
Maximum response length & 2572 \\
Maximum SQL start length traing & 2048 \\
Maximum SQL start length evaluation &  8192  \\
Maximum observation length & 512 \\
Evaluation batch size & 16 by default on 8 GPUs \\
Selection methods & greedy \\
Selection methods & reward\_max \\
Selection methods & consistency \\
Selection methods &  Best-of-N \\
% Optional selector & GenRM if external model is provided \\
\bottomrule
\end{tabular}
\caption{Rollout, decoding, and selection settings.}
\label{tab:appendix_rollout_eval}
\end{table}

\begin{table}[htbp]
\centering
\footnotesize
\setlength{\tabcolsep}{3pt}
\renewcommand{\arraystretch}{1.12}
\begin{tabularx}{\columnwidth}{
@{}
>{\raggedright\arraybackslash}p{0.34\columnwidth}
>{\raggedright\arraybackslash}X
@{}
}
\toprule
\textbf{Component} & \textbf{Setting} \\
\midrule
Reward manager & SQL reward manager \\
Format constraint & Exactly one \texttt{\textless solution\textgreater} span \\
Required reasoning & At least one \texttt{\textless think\textgreater} span \\
Invalid nesting &
No \texttt{\textless think\textgreater}, \texttt{\textless sql\textgreater}, or
\texttt{\textless observation\textgreater} inside
\texttt{\textless solution\textgreater} \\
Observation rule &
Each \texttt{\textless/observation\textgreater} must be followed by
\texttt{\textless think\textgreater} \\
Invalid format reward & $-1$ \\
Correct execution reward & $+1$ \\
Incorrect execution reward & $0$ \\
SQL executor & SQLite execution \\
Result comparison & Exact set match with gold SQL results \\
Execution timeout & 30 seconds \\
Training reward processes & 36 \\
Evaluation reward processes & 8 \\
\bottomrule
\end{tabularx}
\caption{Reward and verification settings.}
\label{tab:appendix_reward_verifier}
\end{table}

\subsection{Prompts}
In this section, we describe the four prompt templates used by SERL-SQL. 
First, the schema grounding prompt identifies the tables and columns required for answering the question by jointly considering the database schema, question, and retrieved value hints, as shown in Figure~\ref{fig:Grounding}. 
Second, the SQL action prompt guides the student agent to generate an exploratory SQL query under the grounded schema, without directly producing the final answer, as shown in Figure~\ref{fig:SQL_Action}. 
Third, the SQL observation prompt conditions the agent on database execution feedback and determines whether to revise the SQL or output the final solution, as shown in Figure~\ref{fig:SQL_Observation}. 
Finally, the execution-hindsight teacher prompt is used only during training to re-score the student's on-policy trajectory with privileged execution feedback, as shown in Figure~\ref{fig:Hindsight}. 
Together, these prompts support schema grounding, interactive SQL exploration, feedback-driven revision, and hindsight-based credit assignment.

\begin{figure}[ht]
  \centering
  \includegraphics[width=\linewidth]{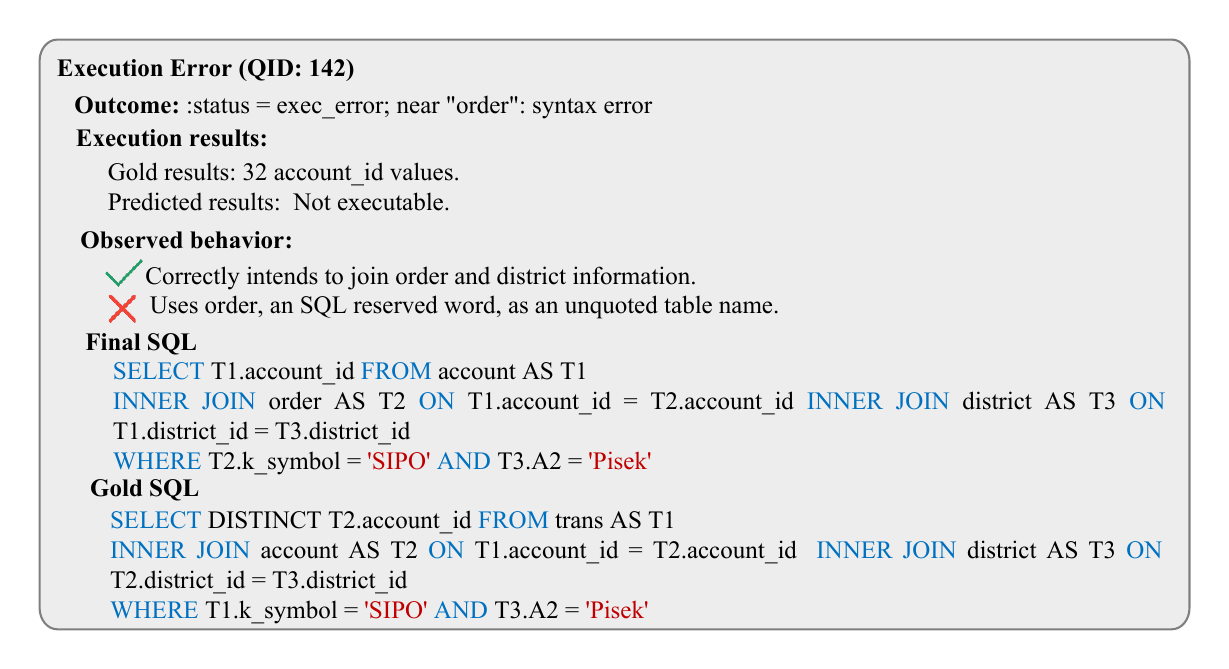}
  \caption{
  Execution error case. The predicted SQL follows a plausible join structure but becomes non-executable because it uses a reserved SQL keyword as an unquoted table name.
  }
  \label{fig:execution_case}
\end{figure}

\begin{figure}[ht]
  \centering
  \includegraphics[width=\linewidth]{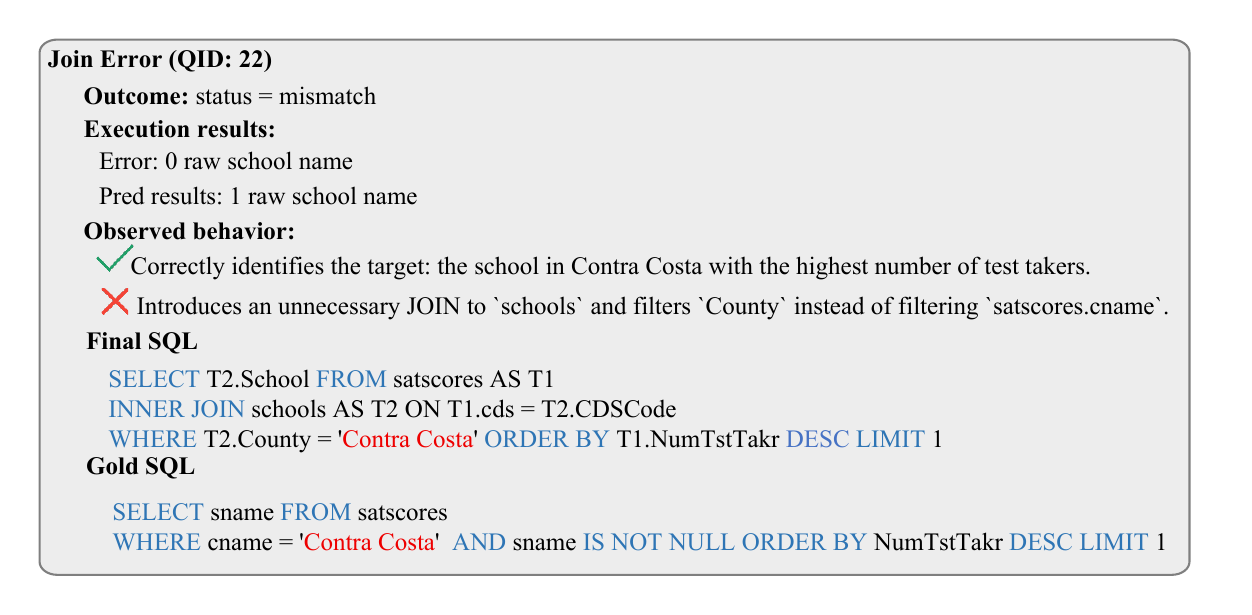}
  \caption{
  Join error case. The model identifies the target entity but introduces an unnecessary join and applies the filtering condition to the wrong table.
  }
  \label{fig:join_case}
\end{figure}

\begin{figure}[ht]
  \centering
  \includegraphics[width=\linewidth]{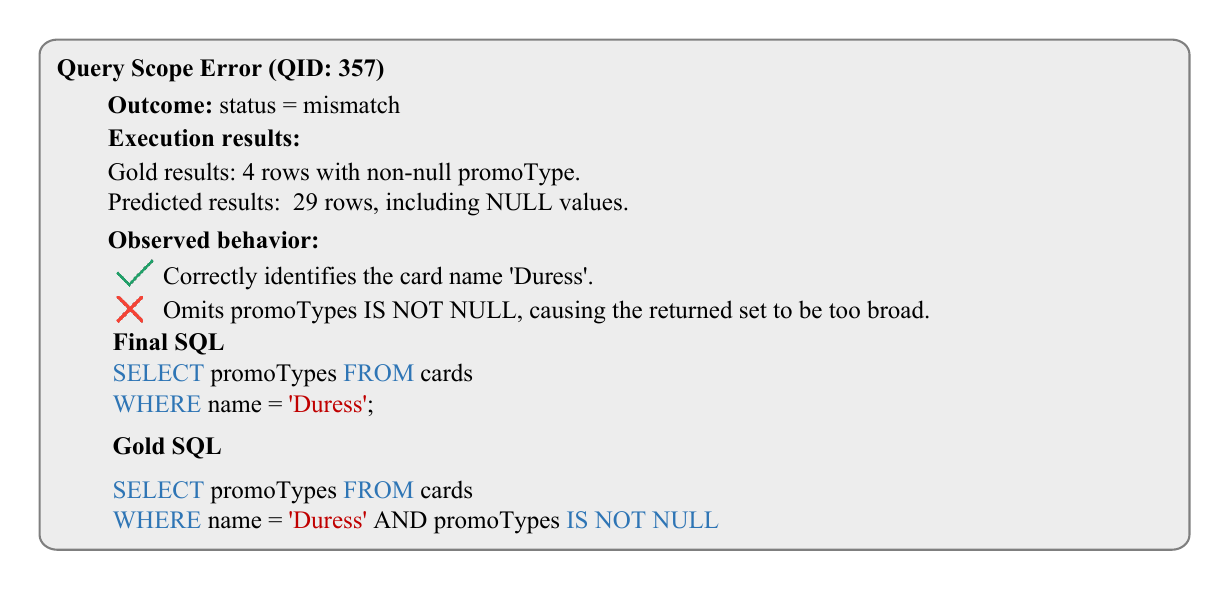}
  \caption{
  Query-scope error case. The predicted SQL retrieves the correct attribute but omits a required non-null constraint, producing an overly broad result set.
  }
  \label{fig:query_scope_case}
\end{figure}

\begin{figure}[ht]
  \centering
  \includegraphics[width=\linewidth]{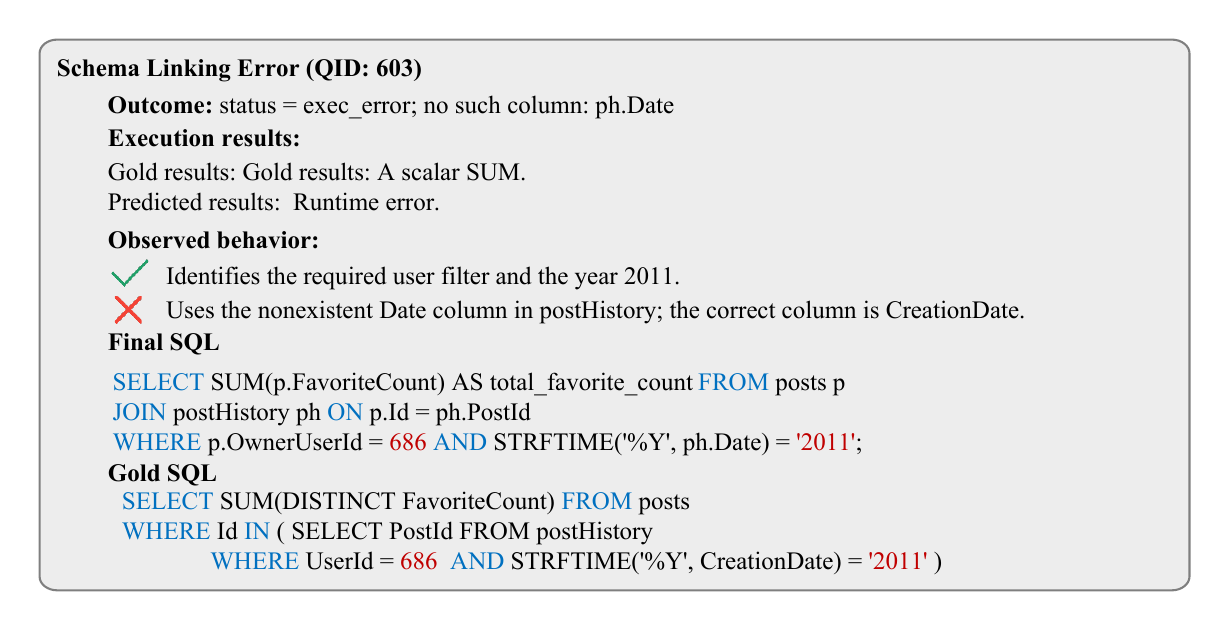}
  \caption{
  Schema-linking error case. The model captures the intended user and temporal condition but links to a nonexistent schema field, causing execution failure.
  }
  \label{fig:schema_case}
\end{figure}

\begin{figure}[ht]
  \centering
  \includegraphics[width=\linewidth]{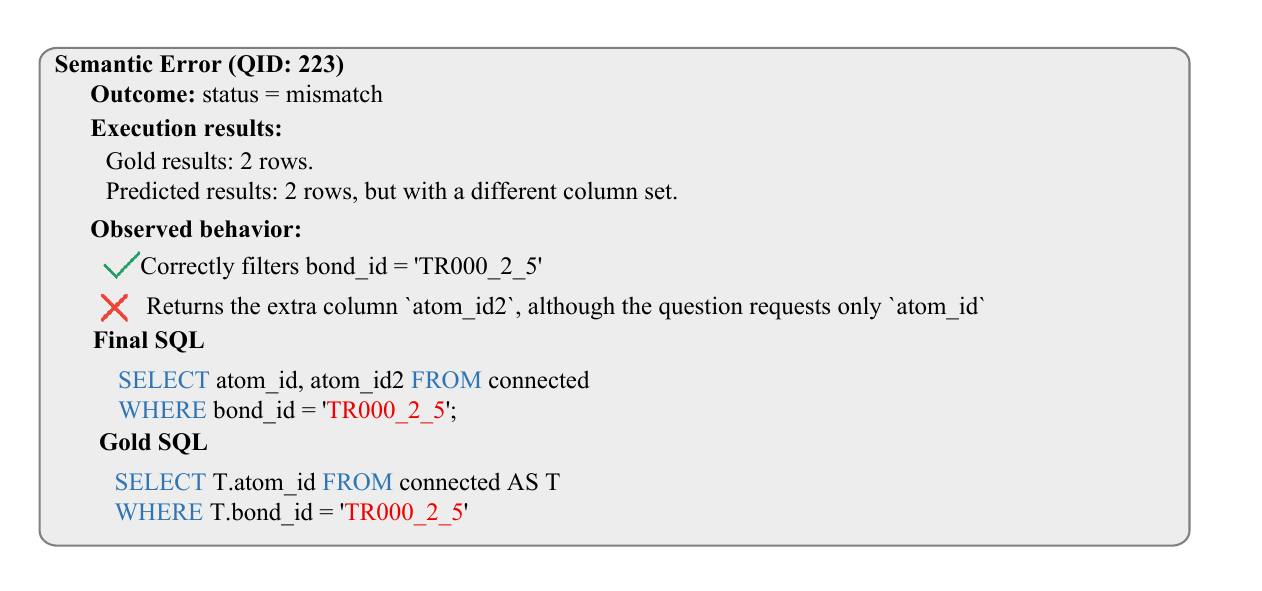}
  \caption{
  Semantic error case. The predicted SQL satisfies the filtering condition but projects an extra column, leading to a result schema inconsistent with the question.
  }
  \label{fig:semantic_case}
\end{figure}

\begin{figure}[ht]
  \centering
  \includegraphics[width=\linewidth]{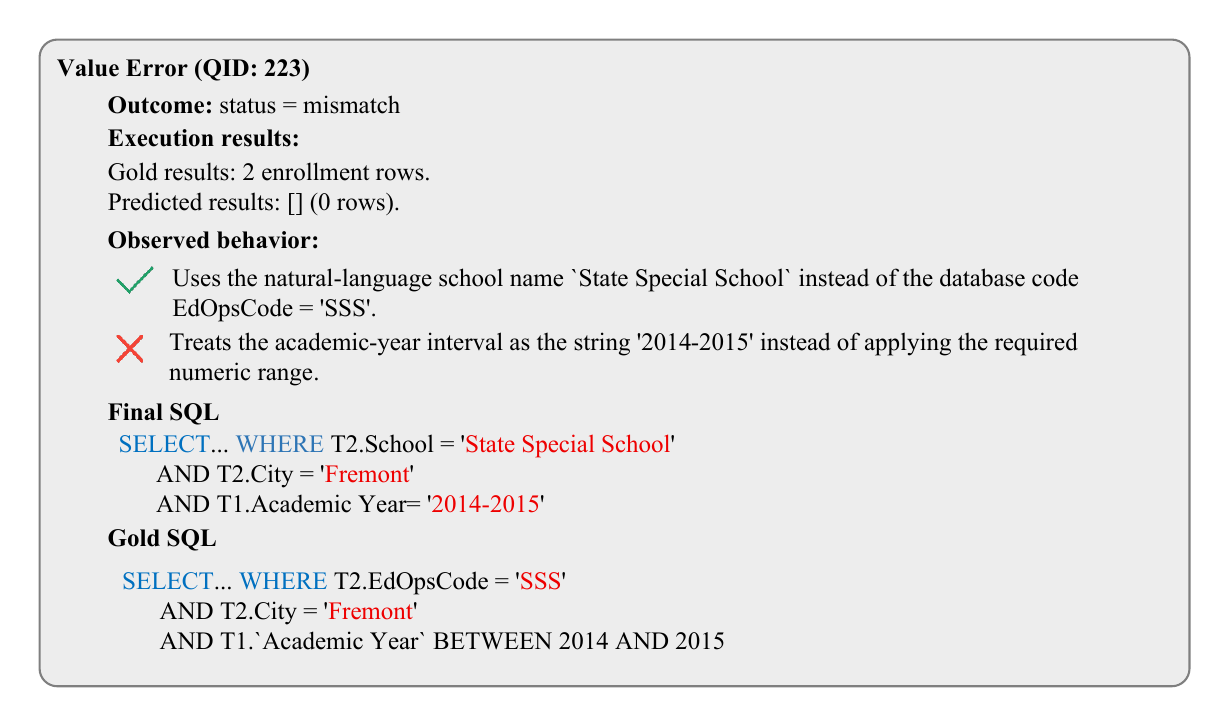}
  \caption{
  Value error case. The model fails to map natural-language values to database-specific codes and mishandles the temporal range, resulting in an empty prediction.
  }
  \label{fig:value_case}
\end{figure}

% \begin{figure}[ht]
%   \centering
%   \includegraphics[width=\linewidth]{Figures/Value.pdf}
%   \caption{
%     Hindsight teacher-signal decay analysis on BIRD-Dev. A moderate decay horizon $K_{\mathrm{decay}}=100$ achieves the best reward\_max EX, outperforming both no-hindsight and non-decayed teacher settings. This indicates that execution hindsight is most useful as transient credit guidance rather than persistent privileged supervision.
%     }
%   \label{fig:Value}
% \end{figure}

\begin{figure*}[h]
  \centering
  \includegraphics[width=0.8\textwidth]{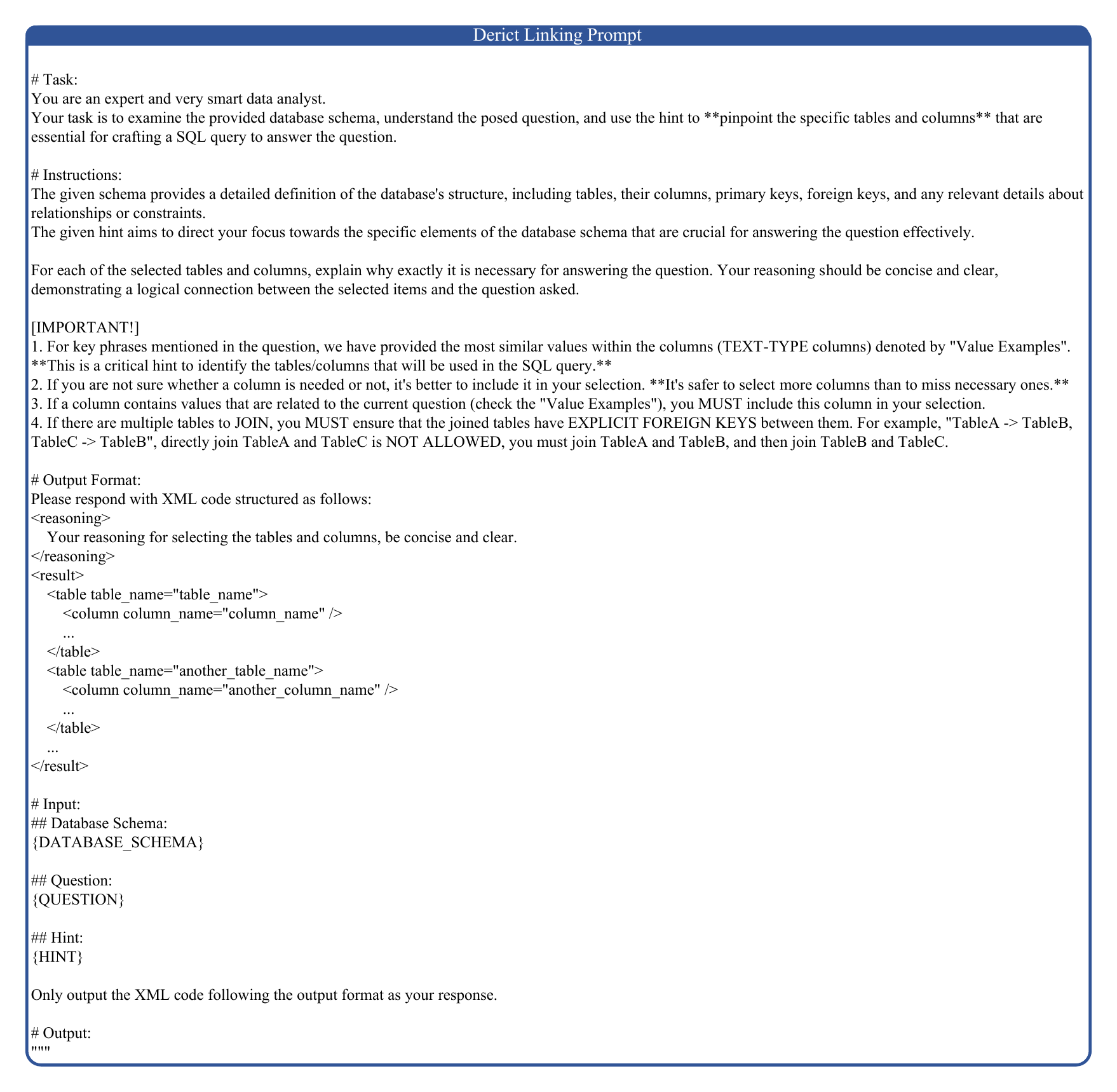}
  \caption{
    Schema grounding prompt. The prompt selects question-relevant tables and columns by combining schema information, the user question, and retrieved value hints.
    }
  \label{fig:Grounding}
\end{figure*}

\begin{figure*}[h]
  \centering
  \includegraphics[width=0.8\textwidth]{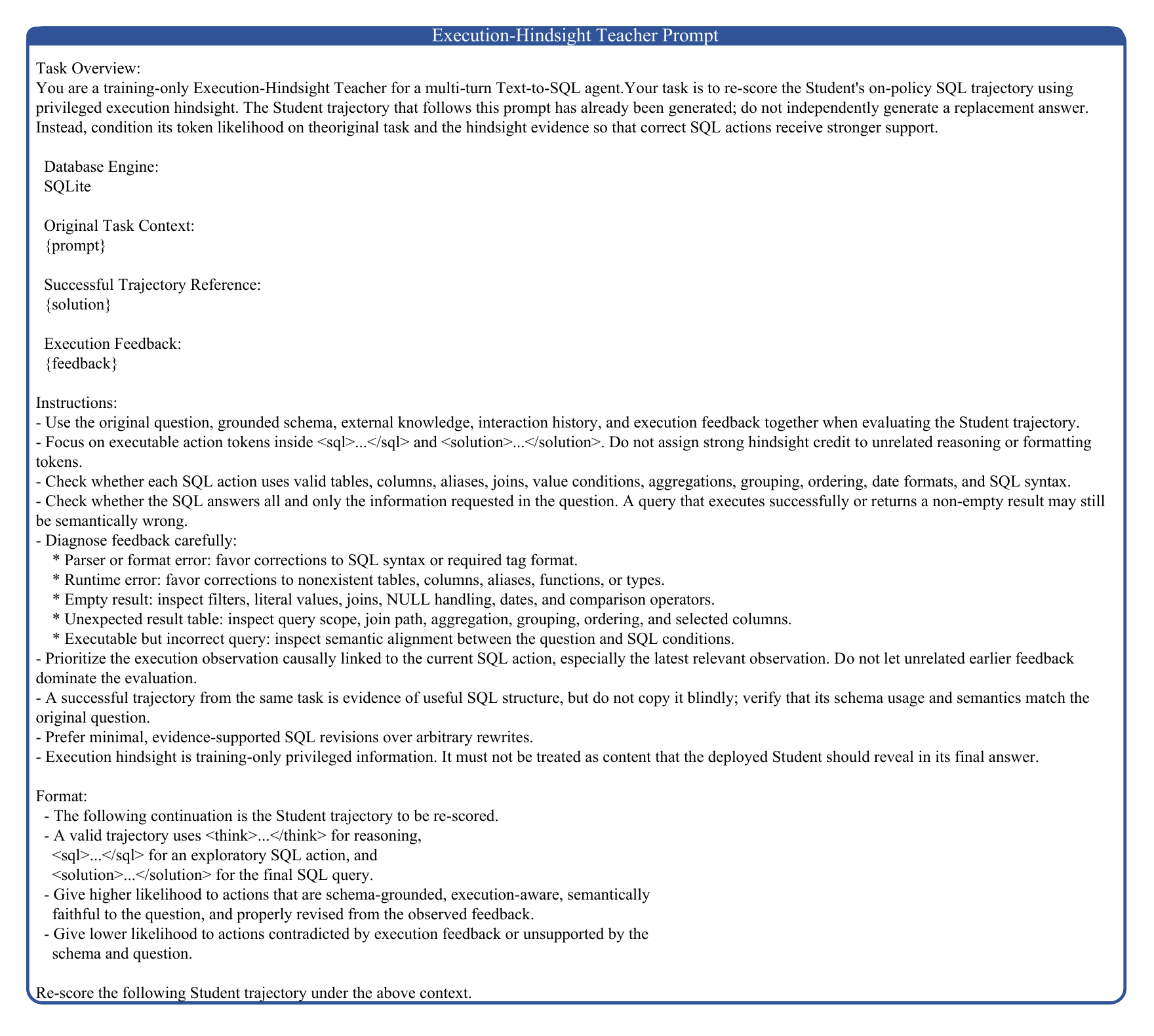}
  \caption{
    Execution-hindsight teacher prompt. The training-only teacher re-scores the student's on-policy SQL trajectory using privileged execution feedback.
    }
  \label{fig:Hindsight}
\end{figure*}

\begin{figure*}[h]
  \centering
  \includegraphics[width=1.0\textwidth]{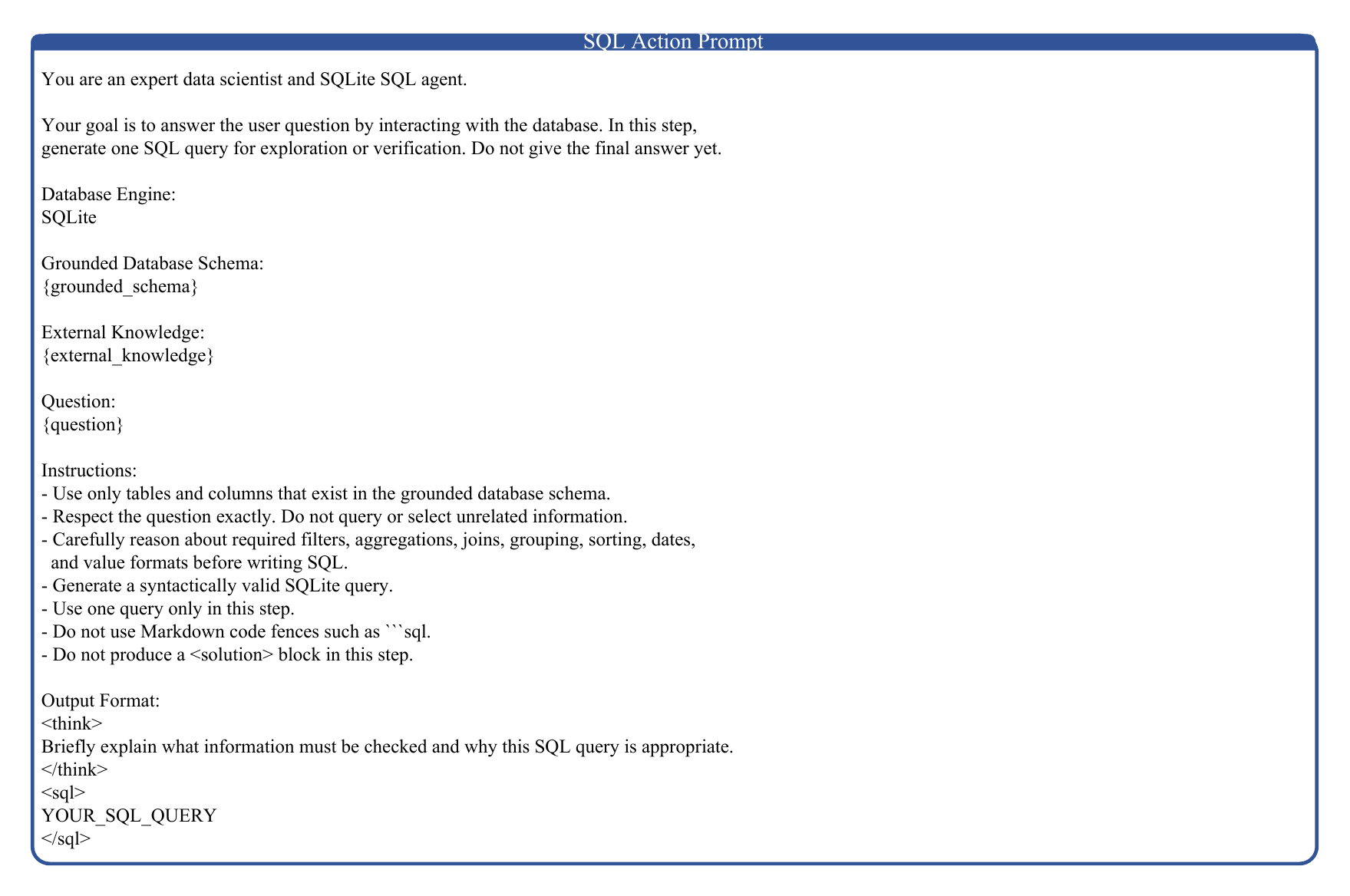}
  \caption{
    SQL action prompt. The student agent generates one exploratory SQL query under the grounded schema for database interaction.
    }
  \label{fig:SQL_Action}
\end{figure*}

\begin{figure*}[h]
  \centering
  \includegraphics[width=1.0\textwidth]{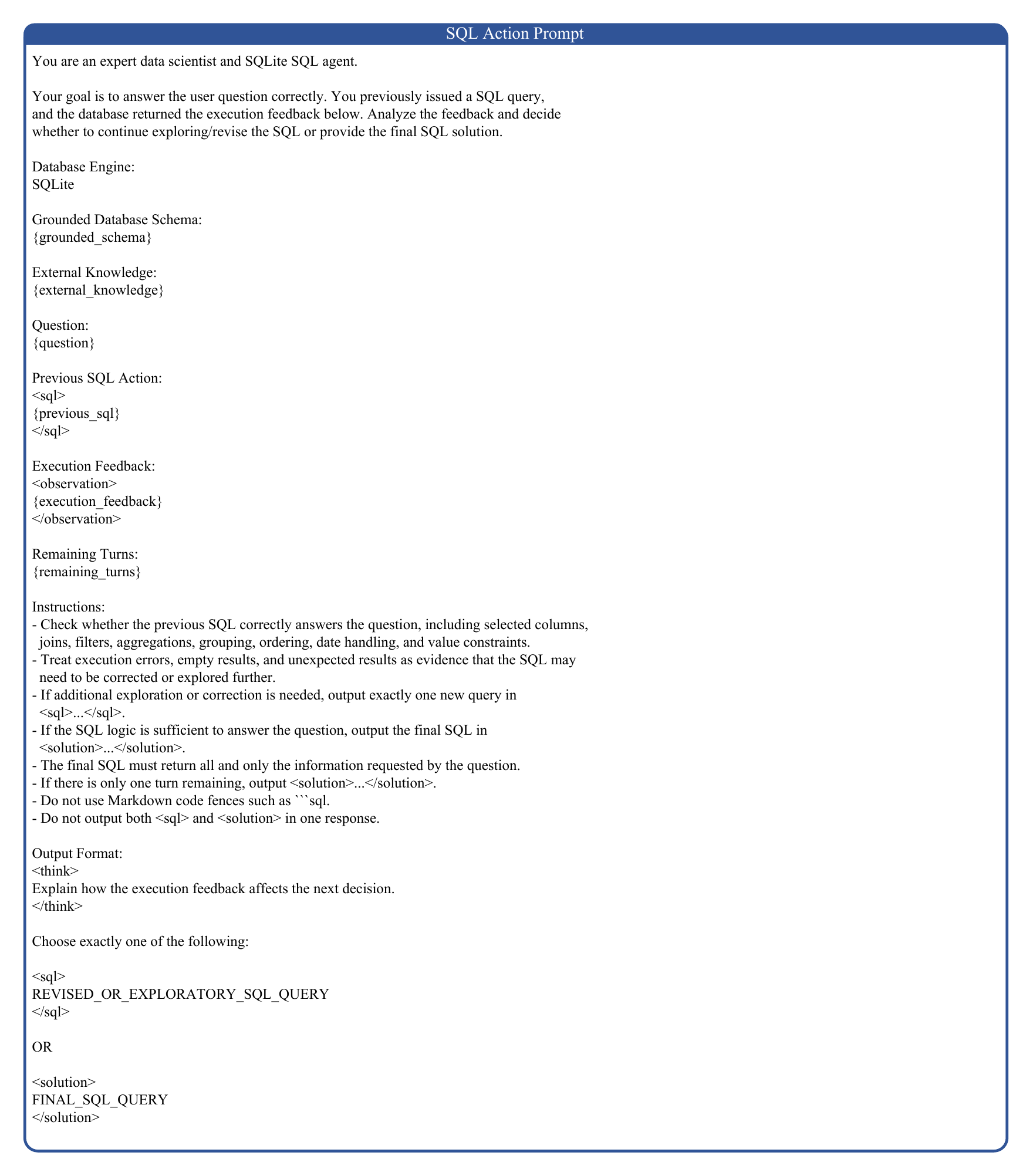}
  \caption{
    SQL observation prompt. The student agent analyzes execution feedback and decides whether to revise the SQL or output the final solution.
    }
  \label{fig:SQL_Observation}
\end{figure*}

\end{document}